\documentclass[lettersize,journal]{IEEEtran}
\usepackage{algorithmic}
\usepackage{algorithm}
\usepackage{array}
\usepackage{textcomp}
\usepackage{stfloats}
\usepackage{url}
\usepackage{verbatim}
\usepackage{graphicx}
\usepackage{cite}

\usepackage{amsmath,amssymb,amsfonts}
\usepackage[font=small]{caption}
\usepackage{lipsum}
\usepackage{enumerate}
\usepackage{amsthm}
\usepackage{color}
\usepackage{threeparttable}
\usepackage{float}
\usepackage{makecell}
\usepackage{booktabs}
\usepackage{listings}
\usepackage{gensymb}
\usepackage{multirow}
\usepackage{color,soul}
\usepackage[]{hyperref}
\usepackage{todonotes}
\usepackage{bm}

\newtheorem{definition}{Definition}
\newtheorem{lemma}{Lemma}
\newtheorem{theorem}{Theorem}
\newtheorem{assumption}{Assumption}
\newtheorem{corollary}{Corollary}

\newtheorem{remark}{Remark}
\usepackage{subfig}

\usepackage{nicefrac}       
\usepackage{microtype}      
\usepackage{xcolor}         
\usepackage{siunitx}

\begin{document}

\title{Conformal Symplectic Optimization for Stable Reinforcement Learning}

\author{
Yao Lyu, 
Xiangteng Zhang,
Shengbo Eben Li*,
Jingliang Duan,
Letian Tao,
Qing Xu,
Lei He,
Keqiang Li
\thanks{This study is supported by National Key R\&D Program of China with 2022YFB2502901, National Natural Science Foundation of China under No. U20A20334, and Tsinghua University-Toyota Joint Research Center for AI Technology of Automated Vehicle. All correspondences should be sent to S. Eben Li.}
\thanks{Y. Lyu, X. Zhang, Q. Xu, and K. Li are with the School of Vehicle and Mobility, Tsinghua University, Beijing, China. Email: (y-lv19@mails.; zhangxt22@mails.; qingxu@; likq@)tsinghua.edu.cn.
}
\thanks{S. Eben Li is with the School of Vehicle and Mobility and the College of Artificial Intelligence, Tsinghua University, Beijing, China. Email: lishbo@tsinghua.edu.cn.
}
\thanks{J. Duan is with the School of Mechanical Engineering, University of Science and Technology Beijing, Beijing, China. Email: duanjl@ustb.edu.cn.
}
\thanks{L.Tao and L. He are with the State Key Laboratory of Intelligent Green Vehicle and Mobility, Tsinghua University, Beijing, China. Email: (tlt22@mails.; helei2023@)tsinghua.edu.cn.
}
}



\maketitle

\begin{abstract}
Training deep reinforcement learning (RL) agents necessitates overcoming the highly unstable nonconvex stochastic optimization inherent in the trial-and-error mechanism. To tackle this challenge, we propose a physics-inspired optimization algorithm called relativistic adaptive gradient descent (RAD), which enhances long-term training stability. By conceptualizing neural network (NN) training as the evolution of a conformal Hamiltonian system, we present a universal framework for transferring long-term stability from conformal symplectic integrators to iterative NN updating rules, where the choice of kinetic energy governs the dynamical properties of resulting optimization algorithms. By utilizing relativistic kinetic energy, RAD incorporates principles from special relativity and limits parameter updates below a finite speed, effectively mitigating abnormal gradient influences. Additionally, RAD models NN optimization as the evolution of a multi-particle system where each trainable parameter acts as an independent particle with an individual adaptive learning rate. We prove RAD's sublinear convergence under general nonconvex settings, where smaller gradient variance and larger batch sizes contribute to tighter convergence. Notably, RAD degrades to the well-known adaptive moment estimation (ADAM) algorithm when its speed coefficient is chosen as one and symplectic factor as a small positive value. Experimental results show RAD outperforming nine baseline optimizers with five RL algorithms across twelve environments, including standard benchmarks and challenging scenarios. Notably, RAD achieves up to a 155.1\% performance improvement over ADAM in Atari games, showcasing its efficacy in stabilizing and accelerating RL training.
\end{abstract}


\section{Introduction}\label{sec_introduction}

Reinforcement learning (RL) stands out as a prominent area within the artificial intelligence community, which promises to provide solutions for decision-making and control of large-scale and complex problems \cite{li2023reinforcement}. In recent years, RL has shown great potential in various challenging domains, including games \cite{silver2016mastering,vinyals2019grandmaster}, robotics \cite{fujimoto2018addressing,duan2021distributional}, autonomous driving \cite{guan2022integrated, duan2024encoding}, etc. However, training an RL agent with neural networks (NNs) is extremely unstable compared to other machine learning tasks since the inherent trial-and-error mechanism brings intractable uncertainty \cite{mnih2015human, duan2024landscape}. Specifically, NNs in RL are prone to overfitting \cite{lillicrap2016continuous}, overestimation \cite{van2016deep,haarnoja2018soft}, and divergence \cite{schulman2015trust}, making it extremely important to guarantee the training stability.

Training an NN involves solving a nonconvex stochastic optimization problem, and gradient-based approaches are commonly employed to address it iteratively, whose stability plays a critical role in the convergence of RL. The history of gradient-based optimizers originated with the stochastic gradient descent (SGD) algorithm \cite{yazan2017comparison}, which estimates the gradient using a small data batch at each step. Hinton et al. \cite{goodfellow2016deep} introduced momentum and developed the stochastic gradient descent with momentum (SGD-M) algorithm, which accelerates convergence and mitigates local minima or saddle points. Nesterov \cite{bottou2018optimization} further incorporates lookahead gradients to update parameters, proposing the Nesterov accelerated gradient (NAG) algorithm, which achieves faster and more stable convergence. Duchi et al. \cite{duchi2011adaptive} presented the adaptive gradient (AdaGrad) algorithm, which dynamically adjusts the learning rate for each parameter based on past gradients, making it suitable for sparse or noisy gradients. Hinton \cite{goodfellow2016deep} developed the root mean square propagation (RMSprop) algorithm, which adapts the learning rate for each parameter using an exponential moving average of squared gradients instead of accumulating all past gradients, thereby preventing excessive drop in the learning rate. Ultimately, Kingma and Ba \cite{kingma2014adam} combined the two algorithms above to propose the adaptive moment estimation (ADAM) algorithm, which is currently one of the most widely used optimizers for NNs. However, previous algorithms often solely rely on empirical and pragmatic approaches to expedite the convergence of NNs, lacking theoretical foundations, particularly in analyzing the optimization dynamics and investigating training stability.

Recent studies underscore the importance of drawing parallels between gradient-based optimization algorithms and continuous-time dynamical systems \cite{wibisono2016variational, francca2021gradient}. Analyzing optimization algorithms through their corresponding dynamical systems allows us to leverage extensive knowledge about these systems. This approach facilitates the examination of crucial properties such as stability and convergence rate that emerge during the optimization process. Furthermore, dissipative dynamical systems are highly desirable for optimization due to their natural convergence towards stationary points \cite{jordan2018dynamical}. Among these systems, conformal Hamiltonian systems stand out, as their declining Hamiltonian (i.e., total energy) and symplectic form (i.e., phase area) over time ensure both convergence and long-term stability \cite{francca2021dissipative}.

Specifically, an optimization algorithm can be viewed as a discretized representation of a dynamical system that preserves essential properties of the continuous-time flow \cite{muehlebach2021optimization}. Conformal symplectic integrators ensure that symplecticity and phase portrait are maintained after discretization, thus preserving the long-term stability and convergence rates of the continuous conformal Hamiltonian system \cite{feng2010symplectic, bhatt2016second}. Well-established algorithms such as heavy-ball (HB) and dissipative leapfrog (DLPF) exemplify this connection \cite{francca2021dissipative}.\footnote{SGD-M can be considered the stochastic counterpart of HB, and both will be collectively referred to as HB in this paper.} However, despite their effectiveness in stabilizing training, these algorithms' performance may degrade with large gradients due to unconstrained parameter updating speeds. Relativistic gradient descent (RGD), originating from a relativistic system, addresses this issue by constraining the updating speed to a finite value \cite{francca2020conformal}. Nevertheless, all these algorithms model NN training as systems governed by a single multi-dimensional particle, adjusting all trainable parameters with equal magnitude. This approach results in limited adaptability and suboptimal convergence for RL \cite{Wilson2017marginal}.

This paper proposes the relativistic adaptive gradient descent (RAD) algorithm based on the conformal symplectic discretization of a relativistic multi-particle system, aiming to stabilize RL training over extended periods. We outline the key contributions as follows:
\begin{enumerate}
\item We pioneer incorporating symplectic preservation into NN optimization to enhance long-term training stability. By formulating NN training as an evolving conformal Hamiltonian system, we develop a universal framework that transfers long-term stability from conformal symplectic integrators to NN updating rules. Specifically, the kinetic energy formulation governs optimization dynamics. The classical kinetic energy results in accelerated gradient methods like HB and DLPF, which permit unconstrained parameter updates. To enhance training stability, we suggest a relativistic formulation that places natural limits on updating speeds, thereby mitigating the destabilizing effects caused by large gradients.
\item The proposed RAD algorithm models NN training as a multi-particle system, considering each trainable parameter as an independent relativistic particle. This perspective allows for individual adaptivity and constrained updating speeds for each parameter. To extend RAD to general stochastic optimization, we integrate exponential moving average and bias correction to handle optimization noise and stochasticity. We also establish a connection between RAD and the well-known ADAM algorithm. While ADAM manually incorporates a small rational factor to prevent division by zero, RAD naturally maintains a growing symplectic factor controlling adaptivity and training stability. This can even be linked to the mass energy of the conformal Hamiltonian system to some extent, shedding light on intrinsic dynamics within other mainstream adaptive gradient methods.
\item We develop a convergence analysis of RAD under general nonconvex stochastic optimization. We prove that RAD exhibits sublinear convergence to a stationary point, where the error bound decays inversely with the number of maximum iterations. Moreover, more minor gradient variance and larger batch sizes contribute to tighter convergence. By grounding RAD in stochastic optimization and proving convergence for general objectives, we expand RAD's potential applications for stabilizing training across machine learning tasks beyond solely RL.
\end{enumerate}

Additionally, we conduct empirical evaluations of RAD on MuJoCo benchmarks \cite{todorov2012mujoco} and Atari games \cite{bellemare2013arcade}, showcasing its ability to stabilize RL training. The results demonstrate that RAD outperforms state-of-the-art optimizers, achieving higher policy performance with faster convergence. These findings underscore the potential for RAD to smooth and accelerate optimization for complex RL tasks.

The remaining part of this paper is organized as follows: Section \ref{section_preliminaries} introduces conformal Hamiltonian systems and conformal symplectic integrators. Section \ref{section_dynamics_NN_training} presents the perspective of NN training from a conformal Hamiltonian viewpoint. Then, we propose RAD in Section \ref{section_rad} and establish its convergence in Section \ref{section_convergence_analysis}. Subsequently, empirical experiments are conducted in Section \ref{section_experiments}, and our work is concluded in Section \ref{section_conclusion}.

\section{Preliminaries}\label{section_preliminaries}
This section briefly introduces the conformal Hamiltonian system and its integrator with the capacity of symplectic preservation. 
\subsection{Conformal Hamiltonian system}
The state of a conformal Hamiltonian system can be determined by a point in the phase space, $(q,p) \in \mathbb{R}^{2n}$, where $q=q(t)\in\mathbb{R}^n$ are the generalized coordinates, $p=p(t)\in\mathbb{R}^n$ are their conjugate momenta, $t \geq 0$ is the time, and $n \in \mathbb{N}_{+}$ is the dimension of the system. the Hamiltonian $H:\mathbb{R}^{2n} \rightarrow \mathbb{R}$ characterizes the total energy of the system:
\begin{equation}
\nonumber
\label{eq_classic_H}
    H(q,p)=T(p)+U(q),
\end{equation}
where $T(p)$ and $U(q)$ represent kinetic and potential energy, respectively. A conformal Hamiltonian system is required to obey the following canonical equations:
\begin{equation}
\label{eq_conformal_H_equation_1}
\dot{p} = -\nabla_q H(q,p)-rp, \quad \dot{q} = \nabla_p H(q,p), 
\end{equation}
where $r>0$ is a damping constant and we can rewrite \eqref{eq_conformal_H_equation_1} in matrix expression as
\begin{equation}
\label{eq_conformal_H_2}
\dot{z}=S \nabla H(z)-r D z,
\end{equation}
where
\begin{displaymath}
z =\left[\begin{array}{l}
q \\
p
\end{array}\right], \quad S=\left[\begin{array}{cc}
0 & I \\
-I & 0
\end{array}\right], \quad D=\left[\begin{array}{ll}
0 & 0 \\
0 & I
\end{array}\right],
\end{displaymath}
and $I$ is the $n$-dimensional identity matrix. Canonical equations \eqref{eq_conformal_H_2} define a continuous flow describing the evolution of the system as $\varphi_t: \varphi_t(z_0)=z(t)$, where $z_0=z(0)$. Moreover, we use $\mathcal{C}(z)=S \nabla H(z)$ and $\mathcal{D}(z)=-r D z$ to denote the ``conservative" and ``dissipative" parts and associate them with the continuous flows $\varphi_t^{\mathcal{C}}$ and $\varphi_t^{\mathcal{D}}$, respectively.

Applying the wedge product induces the symplectic form: $\omega_t={\rm d}q(t)\wedge {\rm d}p(t)$, i.e., phase area. The system flow $\varphi_t$ contracts the phase area exponentially, i.e., $\omega_t = e^{-rt} \omega_0$. As the phase area $\omega_t$ represents a subset of the phase space that the system may visit at time $t$, its compression inhibits unstable behavior of the system state, preventing it from becoming an outlier during system evolution and ultimately leading to convergence towards stationary points\cite{hairer2006geometric}.

\subsection{Conformal symplectic integrator}
To preserve dynamic properties during discretization of the conformal Hamiltonian system, we adopt the splitting method:
\begin{enumerate}
    \item Individually discretize the "conservative" flow $\varphi_t^{\mathcal{C}}$ and "dissipative" flow $\varphi_t^{\mathcal{D}}$;
    \item Composite them to create a numerical map $\phi_h$ that approximates the system flow $\varphi_t$ over a small step size $h$.
\end{enumerate}

We first obtain the numerical map $\phi_{h}^{\mathcal{C}}$ of the conservative part $\dot{z}=\mathcal{C}(z)=S \nabla H(z)$, which approximates $\varphi_{t}^{\mathcal{C}}$ in a small time interval $[t,t+h]$. We can choose any standard symplectic integrator to achieve this goal. For example, if the symplectic Euler method is selected, we have $\phi_{h}^{\mathcal{C}}:(q,p)\rightarrow\left(Q,P\right)$, where $P=p-h \nabla_{q} H\left(q, P\right)$ and $Q=q+h \nabla_{p} H\left(q, P\right)$ \cite{hairer2006geometric}. Then, the dissipative part, i.e., $\dot{z}=\mathcal{D}(z)=-r D z$, can be integrated explicitly as $\phi_{h}^{\mathcal{D}}:(q, p) \rightarrow\left(q, e^{-r h} p\right)$. Finally, with the composition $\phi_{h}=\phi_{h}^{\mathcal{C}} \circ \phi_{h}^{\mathcal{D}}$, we obtain a first-order conformal symplectic integrator $\phi_{h}$ as
\begin{equation}
\label{eq_step3_1}
\begin{aligned}
p_{k+1}&=e^{-r h} p_{k}-h \nabla_{q} H\left(q_{k}, p_{k+1}\right), \\
q_{k+1}&=q_{k}+h \nabla_{p} H\left(q_{k}, p_{k+1}\right),
\end{aligned}
\end{equation}
where $q_k$ represents the value of $q$ at the $k$-th discrete iteration of $\phi_h$. This should approximate the value of $q$ at time $t = kh$ during the continuous evolution $\varphi_t$, i.e., $q_k \approx q(kh)$.

Similarly, if we choose the leapfrog method to approximate $\varphi_t^\mathcal{C}$ and consider the composition $\phi_{h}=\phi_{h / 2}^{\mathcal{D}} \circ \phi_{h}^{\mathcal{C}} \circ \phi_{h / 2}^{\mathcal{D}}$, we immediately obtain a second-order numerical integrator \cite{hairer2006geometric, francca2020conformal}.\footnote{Please refer to the supplementary material, which can be accessed on \href{https://github.com/TobiasLv/RAD}{https://github.com/TobiasLv/RAD} along with the Python code.} Conformal symplectic integrators preserve the declining property of Hamiltonian and symplectic form of the conformal Hamiltonian system with negligible approximating error over significantly extended periods, maintaining convergence rate while ensuring long-term stability \cite{muehlebach2021optimization,francca2021dissipative}.

\section{Conformal Hamiltonian perspective on neural network training}
\label{section_dynamics_NN_training}
This section presents a novel perspective on viewing NN training as the evolution of a conformal Hamiltonian system. We present a universal framework for designing innovative conformal symplectic integrators as iterative updating rules for NNs, where the choice of kinetic energy plays a crucial role in governing the optimization dynamics. Classical kinetic energy enables accelerated gradient algorithms without limitations on parameter updating speed. In contrast, relativistic kinetic energy allows for adaptive gradient methods and imposes constraints on parameter updating speed to mitigate the impact of excessively large gradients.

\subsection{Mapping NN Training to conformal Hamiltonian dynamics}
Training an NN is typically formulated as a general nonconvex stochastic optimization problem:
\begin{equation}
\label{eq_general_optimization}
\min _{\theta \in \mathbb{R}^{n}} J(\theta)=\mathbb{E}_{x \sim \mathcal{P}}\{\mathcal{L}(x, \theta)\},
\end{equation}
where $J$ is the objective function, $\mathcal{L}$ is a nonconvex loss function, $\mathcal{P}$ is the distribution of data $x$, and $\theta\in \mathbb{R}^{n}$ are the trainable parameters. 

While convergence requires that $\nabla J(\theta)=0$ and the updating speed of $\theta$ approaches zero \cite{goodfellow2016deep}, this condition parallels that of a conformal Hamiltonian system converging to a stationary point, where $\nabla U(q)=0$ and momenta $p=0$ \cite{jordan2018dynamical}. Based on this analogy, the trainable parameters $\theta$ of an NN can be seen as analogous to the generalized coordinates $q$ of a Hamiltonian system, with the objective function $J(\theta)$ corresponding to the potential energy $U(q)$, i.e.,
\begin{equation}
\label{eq_analogy_NN_H_system}
\theta = q, \;J(\theta)=U(q), \;H(\theta,p)=T(p)+J(\theta).
\end{equation}

Building upon the analogy between NN training and Hamiltonian dynamics, we present a novel framework to develop variable iterative updating rules for NNs, which involves two key steps:

1) Specify the formulation of the kinetic energy $T(p)$ to construct different Hamiltonians $H$. This allows the embedding of different dynamics;

2) Specify the conformal symplectic integrator to discretely approximate the system flow $\varphi_t$. Different integrators will lead to different updating rules.

This analogy provides a promising avenue for embedding dynamical properties or physical principles into NN optimization. While the integrator primarily relates to the precision of discretization, the kinetic energy $T(p)$ directly shapes the dynamics governing optimization trajectories. Consequently, formulating appropriate kinetic energy becomes crucial in influencing salient algorithm behaviors such as stability and convergence rate. Our primary goal is to develop a new NN optimization algorithm that enhances learning stability and performance by determining a suitable expression for $T(p)$.

\subsection{Accelerated gradient from classical kinetic energy}
The classical kinetic energy $T(p)=\nicefrac{||p||^2}{2m}$ is a common choice for developing NN optimizers. For instance, if we substitute the classical kinetic energy into \eqref{eq_analogy_NN_H_system}, we can directly derive the following expressions from \eqref{eq_step3_1}:
\begin{equation}
\label{eq_HB_origin}
p_{k+1}=e^{-r h} p_{k}-h \nabla J\left(\theta_{k}\right), \quad
\theta_{k+1}=\theta_{k}+\frac{h}{m} p_{k+1}.
\end{equation}

We further introduce a change of variables to make them more familiar within the optimization community:
\begin{equation}
\label{eq_variable_change_1}
\begin{aligned}
v_{k}&=-\frac{1-e^{-r h}}{h} p_{k},\text{ } \alpha=\frac{h^2}{m\left(1-e^{-r h}\right)},\text{ } \beta_1=e^{-r h},
\end{aligned}
\end{equation}
where $v\in\mathbb{R}^{n}$ are the conjugate momenta of trainable parameters, $\alpha>0$ is the learning rate, and $0<\beta_1<1$ is the first-order momentum coefficient. By plugging \eqref{eq_variable_change_1} into \eqref{eq_HB_origin}, we obtain:
\begin{equation}
\label{eq_HB}
v_{k+1}=\beta_1 v_{k}+(1-\beta_1) \nabla J\left(\theta_{k}\right), \quad
\theta_{k+1}=\theta_{k}-\alpha v_{k+1},
\end{equation}
which is precisely the heavy-ball (HB) algorithm (see Algorithm \ref{algorithm_HB}) \cite{goodfellow2016deep}. Additionally, substituting classical kinetic energy into a suitable second-order conformal symplectic integrator easily derives the dissipative leapfrog (DLPF) algorithm \cite{francca2021dissipative}. See supplementary materials for more details. 

As shown in \eqref{eq_HB}, discretizing a classical conformal Hamiltonian system inherently incorporates a momentum term, yielding an optimization algorithm that enables accelerated convergence during NN training compared to vanilla gradient descent. Specifically, this momentum allows for accumulating speed in parameter updates over time, effectively empowering the optimization process to escape saddle points and local minima. Therefore, adopting the conformal Hamiltonian viewpoint supplies a principled physics-based approach for embedding accelerated dynamics within NN updating rules.

\begin{algorithm}
\caption{Heavy-ball (HB) \cite{goodfellow2016deep}}
\label{algorithm_HB}
\renewcommand{\algorithmicrequire}{\textbf{Input:}}
\begin{algorithmic}[1]
\REQUIRE parameters of neural network $\theta_0$ and their conjugate momenta $v_0$, learning rate $\alpha>0$, first-order momentum coefficient $0<\beta_1<1$
\FOR{$k=0$ {\bf to} $N-1$}
    \STATE $v_{k+1}=\beta_1 v_{k}+(1-\beta_1) \nabla J\left(\theta_{k}\right)$
    \STATE $g_{k}=v_{k+1}$
    \STATE $\theta_{k+1}=\theta_{k}-\alpha g_{k}$
\ENDFOR
\end{algorithmic}
\end{algorithm}

\subsection{Adaptive gradient from relativistic kinetic energy}
Classical kinetic energy commonly relates to Newtonian systems where the speed has no limitation. In the context of \eqref{eq_HB_origin}, if gradients $\nabla J$ are large enough, they can cause significant increases in momenta $p$. Without constraints on the updating speed of $\theta$, this can result in uncontrolled training behaviors, analogous to the divergence of physical particle positions without speed constraints. To enhance stability and prevent such divergence, it is crucial to introduce constraints on the updating speed of $\theta$. A promising approach to achieve this is by leveraging physical principles.

In special relativity, no particle can travel faster than the speed of light. This fundamental physical law inspires improving the training stability by bounding the updating speed of $\theta$. Discretizing a relativistic Hamiltonian system applies the constraints of special relativity, including the universal speed limit, to parameter updates over iterations. By capping the ``speed" at which parameters can move per iteration, their accumulation is controlled even for large momenta. Specifically, the kinetic energy of a system governed by an $n$-dimensional relativistic particle is denoted as
\begin{equation}
\nonumber
\label{eq_relativistic_T_single_particle}
T(p)= c \sqrt{||p||^{2}+m^{2} c^{2}},
\end{equation}
where $c > 0$ is the speed of light and $m > 0$ is the particle mass. By substituting it into the canonical equations \eqref{eq_conformal_H_equation_1}, we obtain:
\begin{displaymath}
\dot{p}=-\nabla J(\theta)-r p, \quad \dot{\theta}=\frac{c p}{\sqrt{||p||^2+m^{2} c^{2}}}.
\end{displaymath}

Critically, the updating speed $\dot{\theta}$ is inherently bounded due to the relativistic formulation, even for arbitrarily high momenta $p$. This is because $p$ is normalized by the square root term in the denominator, constraining the maximum speed to $c$. Therefore, applying relativistic dynamics enforces a universal speed limit on the parameters, effectively regulating the movement of $\theta$ irrespective of gradient magnitude. This mechanism restrains the unbounded accumulation of parameters, enhancing training stability and preventing divergence.

If we discretize the above equations following \eqref{eq_step3_1} and define the speed coefficient $\delta>0$ as
\begin{equation}
\label{eq_variable_change_2}
\delta=\frac{h}{c m\left(1-e^{-r h}\right)},
\end{equation}
with the change of variables \eqref{eq_variable_change_1}, we can obtain the following iterative updating rules:
\begin{equation}
\label{eq_rgd}
\begin{aligned}
&v_{k+1}=\beta_1 v_{k}+(1-\beta_1) \nabla J\left(\theta_{k}\right), \\
&\theta_{k+1}=\theta_{k}-\frac{\alpha}{\sqrt{\delta^{2} ||v_{k+1}||^{2}+1}} v_{k+1},
\end{aligned}
\end{equation}
which is similar to the relativistic gradient descent (RGD) algorithm (see Algorithm \ref{algorithm_rgd}) \cite{francca2020conformal}. 

Upon examining \eqref{eq_rgd}, the relativistic basis endows the resulting optimization algorithm with not only constrained parameter updates but also the capability to adapt the learning rate based on previous gradients dynamically. Precisely, the learning rate is autonomously adjusted during training, decreasing in response to excessively large momenta and increasing when encountering minute momenta. This reduces the overhead for fine-tuning the learning rate.

\begin{algorithm}
\caption{Relativistic gradient descent (RGD) \cite{francca2020conformal}}
\label{algorithm_rgd}
\renewcommand{\algorithmicrequire}{\textbf{Input:}}
\begin{algorithmic}[1]
\REQUIRE parameters of neural network $\theta_0$ and their conjugate momenta $v_0$, learning rate $\alpha>0$, first-order momentum coefficient $0<\beta_1<1$, speed coefficient $\delta>0$
\FOR{$k=0$ {\bf to} $N-1$}
    \STATE $v_{k+1}=\beta_1 v_{k}+(1-\beta_1) \nabla J\left(\theta_{k}\right)$
    \STATE $g_{k}=v_{k+1}$
    \STATE $\alpha_{k}=\frac{1}{\sqrt{\delta^{2} ||v_{k+1}||^2+1}} \alpha$
    \STATE $\theta_{k+1}=\theta_{k}-\alpha_{k} g_{k}$
\ENDFOR
\end{algorithmic}
\end{algorithm}

\section{Relativistic adaptive gradient descent}
\label{section_rad}
This section proposes the relativistic adaptive gradient descent (RAD) algorithm, which models a relativistic system consisting of multiple independent one-dimensional particles. RAD enables individual adaptivity for each trainable parameter, enhancing its efficiency on complex optimization problems like reinforcement learning (RL). Furthermore, to broaden RAD's applicability beyond its origin from deterministic dynamics, we incorporate exponential moving average and bias correction techniques, making it suitable for general nonconvex stochastic optimization problems. Finally, we discuss the close relationship between RAD and the adaptive moment estimation (ADAM) algorithm, shedding light on studying the inherent dynamics of other adaptive gradient methods.


\subsection{Individual adaptivity from multi-particle systems}
In Section \ref{section_dynamics_NN_training}, we associate NNs with a system governed by a single $n$-dimensional particle, wherein all trainable parameters are treated as separate dimensions of the same particle. This results in limited individual adaptivity. As indicated by \eqref{eq_rgd}, the learning rate of each parameter $\theta_i$ in $\theta = [\theta_1, \theta_2, \cdots, \theta_i, \cdots, \theta_n]^\top$ is uniformly controlled by the denominator $\sqrt{\delta^2||v||^2+1}$ without individual adjustments. This deficiency results in slow convergence towards local optima during nonconvex stochastic optimization like RL.

To address the limitations of a single-particle system, we present an alternative perspective regarding NN optimization as the evolution of a system consisting of multiple independent one-dimensional particles. Under this view, each parameter $\theta_i$ can be treated as a separate one-dimensional particle with position $q_i$ and momentum $p_i$. The optimization process for each trainable parameter then obeys the dynamics of its associated particle. Specifically, a system of equal-mass one-dimensional relativistic particles is described by the following Hamiltonian:
\begin{equation}
\label{eq_relativistic_H}
H(\theta, p)=\sum_{i=1}^{n} c \sqrt{p_{i}^{2}+m^{2} c^{2}}+J(\theta),
\end{equation}
where the kinetic energy is composed of the individual kinetic energy of each particle $i$. Substitute this into \eqref{eq_step3_1}, we can obtain the following updating rules for all $i\in\{1,2,\cdots,n\}$:
\begin{displaymath}
\begin{aligned}
p_{k+1, i}&=e^{-r h} p_{k, i}-h \left[\nabla J\left(\theta_{k}\right)\right]_{i}, \\
\theta_{k+1,i}&=\theta_{k,i}+\frac{hc}{\sqrt{p_{k+1,i}^{2}+m^{2} c^{2}}} p_{k+1,i},\\
\end{aligned}
\end{displaymath}
where the subscript $i$ represents the $i$-th element of the corresponding vector. 

Notably, the updating speed of each parameter ${\theta}_i$ is adapted based on the individual momentum $p_i$. This feature endows NN training with per-parameter adaptivity, facilitating convergence under nonconvex stochastic optimization. By introducing the change of variables \eqref{eq_variable_change_1} and \eqref{eq_variable_change_2}, we immediately receive the original updating rules of the first-order RAD:
\begin{equation}
\label{eq_rad_1_original}
\begin{aligned}
&v_{k+1, i}=\beta_1 v_{k, i}+(1-\beta_1) \left[\nabla J\left(\theta_{k}\right)\right]_{i}, \\
&\theta_{k+1,i}=\theta_{k,i}-\frac{\alpha}{\sqrt{\delta^{2} v_{k+1,i}^{2}+1}} v_{k+1,i}.
\end{aligned}
\end{equation}

Since we model a multi-particle relativistic system, the learning rate of each parameter can be individually adjusted in terms of its learned second-order momentum $v^2_i$. Thus, each parameter can adaptively be updated at a different rate, which helps RAD converge fast under stochastic nonconvex settings.\footnote{In the following content, all operations on vectors are performed element-wise without using the subscript $i$ for brevity.}

Moreover, the speed coefficient $\delta$ is one critical hyperparameter that controls the strength of gradient normalization and the level of adaptivity, influencing the stability and convergence rate of RAD. According to \eqref{eq_rad_1_original}, $|\theta_{k+1, i}-\theta_{k, i}| \leq \alpha / \delta$ is always true. Thus, the parameter updates are restricted as we primarily expected. The upper bound is inversely proportional to $\delta$ or is proportional to $c$. The updating speed becomes unbounded when $\delta \rightarrow 0$, i.e., $c \rightarrow \infty$, like in HB and DLPF, where the optimization process is susceptible to abnormally large gradients. Additionally, a second-order RAD arises when employing a second-order integrator, where relevant derivations can be found in the supplementary materials.

\subsection{Exponential moving average and bias correction}
In the context of stochastic optimization, it is common to utilize the first-order momenta $v$ and the second-order momenta $v^2$ to estimate the gradients' mean $\mathbb{E}\{\nabla J\}$ and secondary raw moment $\mathbb{E}\{\left(\nabla J\right)^{2}\}$, respectively. Empirically, like the updating rule of $v$, the exponential moving average technique benefits stability \cite{zaheer2018adaptive}. Therefore, we can specify that the second-order momenta follows a similar updating rule as 
$$y_{k+1}=\beta_2 y_k+(1-\beta_2)\left(\nabla J(\theta_k)\right)^2,$$ where $y\in\mathbb{R}^{n}$ is the second-order momenta and $0<\beta_2<1$ is the second-order momentum coefficient. 

Furthermore, initializing $v$ and $y$ as both zero is a common strategy of parameter initialization when training NNs. However, zero initialization introduces a significant estimation bias at the early stage of training \cite{kingma2014adam}. Correcting this initialization bias is thus crucial to stabilize the training process:
\begin{displaymath}
\begin{aligned}
\mathbb{E}\left\{\nabla J\left(\theta_{k}\right)\right\}&= \mathbb{E}\left\{v_{k+1}\right\} / \left(1-\beta_{1}^{k+1}\right), \\
\mathbb{E}\left\{\left(\nabla J\left(\theta_{k}\right)\right)^{2}\right\}&=\mathbb{E}\left\{y_{k+1}\right\} / \left(1-\beta_{2}^{k+1}\right) .
\end{aligned}
\end{displaymath}

By integrating the exponential moving average and bias correction techniques, we obtain the improved first-order RAD:
\begin{equation}
\label{eq_rad_1_improved}
\begin{aligned}
v_{k+1}&=\beta_{1} v_{k}+\left(1-\beta_{1}\right) \nabla J\left(\theta_{k}\right), \\
y_{k+1}&=\beta_{2} y_{k}+\left(1-\beta_{2}\right)\left(\nabla J\left(\theta_{k}\right)\right)^{2}, \\
\theta_{k+1}&=\theta_{k}- \underbrace{\frac{\sqrt{1-\beta_{2}^{k+1}}}{\sqrt{{\delta^2 y_{k+1}}+{\zeta_k}}}\alpha}_{\alpha_k}\cdot\frac{v_{k+1}}{1-\beta_{1}^{k+1}}.
\end{aligned}
\end{equation}

Here, we call $\zeta_k=1-\beta_{2}^{k+1}$ the symplectic factor, which gradually increases from 0 to 1 as the training progresses. The noteworthy point is that the effective learning rate $\alpha_k$ is always smaller than $\alpha$ when following the original annealing procedure of $\zeta$. However, a large learning rate shown in the premature optimization stage is beneficial for trainable parameters rapidly reaching the near-optimal space \cite{zaheer2018adaptive}. Therefore, to enable $\alpha_k$ to be large enough at the beginning, we can artificially design another annealing procedure of $\zeta$ where it is firstly set extremely close to 0 and then gradually returns to $1-\beta_2^{k+1}$ with the training process going on. Thus, the final version of RAD is shown in Algorithm \ref{algorithm_rad_final}. In the premature optimization stage, RAD is not completely symplectic since the scheme of $\zeta$ has been changed for fast convergence. However, with $\zeta$ returns to $1-\beta_2^{k+1}$, the symplecticity of RAD is recovered, and the long-term stability is preserved.

\begin{algorithm}
\caption{Relativistic adaptive gradient descent (RAD)}
\label{algorithm_rad_final}
\renewcommand{\algorithmicrequire}{\textbf{Input:}}
\begin{algorithmic}[1]
\REQUIRE parameters of neural network $\theta_0$, first-order momenta $v_0$, second-order-momenta $y_0$, learning rate $\alpha>0$, first-order momentum coefficient $0<\beta_1<1$, second-order momentum coefficient $0<\beta_2<1$, speed coefficient $\delta>0$, monotonically increasing sequence $\left\{\epsilon_k>0\right\}$
\FOR{$k=0$ {\bf to} $N-1$}
    \STATE $v_{k+1}=\beta_1 v_{k}+(1-\beta_1) \nabla J\left(\theta_{k}\right)$
    \STATE $y_{k+1}=\beta_{2} y_{k}+\left(1-\beta_{2}\right)\left(\nabla J\left(\theta_{k}\right)\right)^{2}$
    \STATE $g_{k}=\frac{1}{1-\beta_{1}^{k+1}} v_{k+1}$
    \STATE $\alpha_{k}= \frac{\sqrt{1-\beta_{2}^{k+1}}}{\sqrt{\delta^{2} y_{k+1}+\zeta_k}}\alpha$, where $\zeta_k= \min\{\epsilon_k, 1-\beta_{2}^{k+1}\}$
    \STATE $\theta_{k+1}=\theta_{k}-\alpha_{k} g_{k}$
\ENDFOR
\end{algorithmic}
\end{algorithm}

\begin{remark}
While there exists some resemblance between RAD and RGD, we contend that apart from their shared foundation in special relativity, RAD manifests notable distinctions from RGD in two pivotal aspects: 1) RGD is limited to modeling single-particle systems, which restricts its ability to adjust effective learning rates individually and often leads to slower convergence in the context of RL; 2) While RGD solely relies on deterministic dynamics, RAD is designed explicitly for nonconvex stochastic optimization where the annealing of the symplectic factor plays a crucial role. These arguments will be demonstrated in our numerical experiments.
\end{remark}

\subsection{Correlation with ADAM and some insights} 
Remarkably, if we set $\delta=1$ and $\zeta$ be a fixed rational factor $\epsilon$, the first-order RAD is equivalent to ADAM (see Algorithm \ref{algorithm_adam}), a renowned adaptive gradient method that has shown state-of-the-art performance in various RL tasks \cite{kingma2014adam}. Consequently, the annealing procedure of $\zeta$ allows us to perceive RAD in a two-stage optimization process: 1) initially optimizing like ADAM to quickly reach the near-optimal parameter space; 2) subsequently transitioning to optimize in a conformal symplectic manner for long-term training stability. A similar transition is observed in SWATS \cite{keskar2017improving}, wherein it conditionally switches from ADAM to SGD. However, it is worth noting that SGD lacks symplecticity, and the transfer demonstrated in RAD appears to be more natural and seamless.

This connection provides insights into ADAM's dynamics from the standpoint of dissipative relativistic systems. Notably, the small rational factor $\epsilon=1\times10^{-16}$ in ADAM, originally intended to address numerical issues, coincides with RAD's symplectic factor $\zeta$. However, beyond the speed coefficient $\delta$, we argue that $\zeta$ is another crucial factor controlling gradient normalization and adaptivity. Specifically, $\zeta$ traces back to the term $m^2c^2$ of the relativistic Hamiltonian \eqref{eq_relativistic_H}, reflecting the system's mass energy to some extent. This finding helps explain why limiting the adaptivity of ADAM to a certain degree almost always improves its performance \cite{zaheer2018adaptive}, i.e., choosing an appropriately large rational factor $\epsilon$ enables the algorithm to maintain the dynamical system's properties more accurately, which is an inherent attribute of RAD with an increasing symplectic factor $\zeta$.

\begin{algorithm}
\caption{Adaptive moment estimation (ADAM) \cite{kingma2014adam}}
\label{algorithm_adam}
\renewcommand{\algorithmicrequire}{\textbf{Input:}}
\begin{algorithmic}[1]
\REQUIRE parameters of neural network $\theta_0$, first-order momenta $v_0$, second-order momenta $y_0$, learning rate $\alpha>0$, first-order momentum coefficient $0<\beta_1<1$, second-order momentum coefficient $0<\beta_2<1$, rational factor $\epsilon=1\times10^{-16}$
\FOR{$k=0$ {\bf to} $N-1$}
    \STATE $v_{k+1}=\beta_{1} v_{k}+\left(1-\beta_{1}\right) \nabla J\left(\theta_{k}\right)$
    \STATE $y_{k+1}=\beta_{2} y_{k}+\left(1-\beta_{2}\right)\left(\nabla J\left(\theta_{k}\right)\right)^{2}$ \STATE $g_{k}=\frac{1}{1-\beta_{1}^{k+1}} v_{k+1}$
    \STATE $\alpha_{k} = \frac{\sqrt{1-\beta_{2}^{k+1}} }{\sqrt{y_{k+1}+\epsilon}}\alpha$
    \STATE $\theta_{k+1}=\theta_{k}-\alpha_{k} g_{k}$
\ENDFOR
\end{algorithmic}
\end{algorithm}

This viewpoint further provides a promising approach to researching other adaptive gradient algorithms from the perspective of dynamical systems, like AdaGrad \cite{duchi2011adaptive}, NADAM \cite{dozat2016incorporating}, SWATS \cite{keskar2017improving}, ADAMW \cite{loshchilov2019decoupled}, etc. We believe it is possible to design other optimization algorithms by first choosing a reasonable relativistic conformal Hamiltonian system and then approximating it with a suitable conformal symplectic integrator.

\section{Convergence analysis of RAD}
\label{section_convergence_analysis}
In this section, we will explore the convergence characteristics of RAD, focusing on how it performs in general nonconvex stochastic optimization as outlined in \eqref{eq_general_optimization}. We first state some assumptions as the fundamental basis for conducting convergence analysis. 

\begin{assumption}
\label{assumption_1}
The loss function $\mathcal{L}$ is L-smooth, i.e., $\exists L > 0, \forall \theta_{1}, \theta_{2} \text { and } x$, we have
\begin{equation}
\nonumber
\label{eq_assumption_1}
\left\|\nabla \mathcal{L}\left(x, \theta_{1}\right)-\nabla \mathcal{L}\left(x, \theta_{2}\right)\right\| \leq L\left\|\theta_{1}-\theta_{2}\right\|.
\end{equation}
\end{assumption}

\begin{assumption}
\label{assumption_2}
The loss function $\mathcal{L}$ has gradient bounded on each coordinate, i.e., $\exists M > 0, \forall x, \theta \text{ and } i$, we have
\begin{equation}
\nonumber
\label{eq_assumption_2}
\lvert\left[\nabla \mathcal{L}\left(x, \theta\right)\right]_i\rvert \leq M.
\end{equation}
\end{assumption}

\begin{assumption}
\label{assumption_3}
The variance of each coordinate of the stochastic gradient is bounded, i.e., $\exists \sigma_i > 0, \forall \theta \text{ and } i$, we have
\begin{equation}
\nonumber
\label{eq_assumption_3}
\mathbb{E}_{x \sim \mathcal{P}}\left\{\left(\left[\nabla \mathcal{L}(x, \theta)\right]_{i}-\left[\nabla J(\theta)\right]_{i}\right)^{2}\right\} \leq \sigma_{i}^{2}.
\end{equation}
\end{assumption}

Given these assumptions, we can readily derive the following corollaries.\footnote{Please refer to the appendix for comprehensive proofs of all corollaries, lemmas, and theorems.}
\begin{corollary}
\label{corollary_3}
The objective function $J$ is L-smooth, such that
\begin{equation}
\nonumber
\label{eq_corollary_3}
\left\|\nabla J\left(\theta_{1}\right)-\nabla J\left(\theta_{2}\right)\right\| \leq L\left\|\theta_{1}-\theta_{2}\right\|, \exists L > 0, \forall \theta_{1}, \theta_{2}.
\end{equation}
\end{corollary}

\begin{corollary}
\label{corollary_4}
Each coordinate of the gradient of the objective function $J$ is bounded, such that
\begin{equation}
\nonumber
\label{eq_corollary_4}
\lvert[\nabla J(\theta)]_{i}\rvert \leq M, \exists M > 0, \forall \theta, i.
\end{equation}
\end{corollary}

\begin{corollary}
\label{corollary_5}
The variance of the stochastic gradient is bounded, such that
\begin{equation}
\nonumber
\label{eq_corollary_5}
\mathbb{E}_{x \sim \mathcal{P}}\left\{\|\nabla \mathcal{L}(x, \theta)-\nabla J(\theta)\|^{2}\right\} \leq \sum_{i=1}^{n} \sigma_{i}^{2}=\sigma^{2}, \forall \theta.
\end{equation}
\end{corollary}

Furthermore, we present a lemma to facilitate the convergence analysis of RAD.
\begin{lemma}
\label{lemma_1}
For any $\theta_k\in\mathbb{R}^{n}$, we have the following inequation:
\begin{equation}
\nonumber
\label{eq_lemma_1}
\mathbb{E}\left\{g_{k, i}^{2} \mid \theta_{k}\right\} \leq \frac{\sigma_{i}^{2}}{B_{k}}+\left[\nabla J\left(\theta_{k}\right)\right]_{i}^{2}, \forall k, i,
\end{equation}
where $g_{k,i}$ is the $i$-th coordinate of the stochastic gradient $g_k$ calculated at the $k$-th iteration by using the mini-batch $\mathcal{B}_k$ of which size is $B_k$.
\end{lemma}

Based on the above corollaries and lemma, we ultimately establish the convergence of RAD. For simplicity, our derivation omits the bias correction technique, sets $\beta_1$ to 0, and focuses on the first-order version of RAD. However, it should be emphasized that our analysis can also be extended to include the general case and the second-order version of RAD.
\begin{theorem}
\label{theorem_bound}
Let $\left\{\zeta_k \mid 0<\zeta_k < 1, k=0,1, \cdots, N-1\right\}$ be a monotonically increasing sequence. Assume that $\zeta_0$, $\beta_2$, and $\alpha$ are chosen such that the following conditions hold: $\alpha \leq \frac{\sqrt{\zeta_{0}}}{2 L}$ and $\beta_{2} \geq 1-\frac{\zeta_{0}}{16 M^{2} \delta^{2}}$. Then, for $\theta_k$ generated by using RAD (Algorithm \ref{algorithm_rad_final}) with batch size sequence $\left \{ B_k \mid B_k\in\mathbb{N}_+,k=0,1,\cdots,N-1 \right \}$, we have the following bound:
\begin{equation}
\label{eq_theorem_bound}
\begin{aligned}
\frac{1}{N} &\sum_{k=0}^{N-1} \mathbb{E}\left\{\left\|\nabla J\left(\theta_{k}\right)\right\|^{2}\right\} \\
&\leq O\left(\frac{J\left(\theta_{0}\right)-J\left(\theta^{*}\right)}{\alpha N}+\frac{\sigma^{2}}{2 N} \sum_{k=0}^{N-1} \frac{1}{B_{k} \sqrt{\zeta_k}}\right),
\end{aligned}
\end{equation}
where $\theta^*$ is an optimal solution of problem \eqref{eq_general_optimization}, and $N$ is the number of maximum iterations. 
\end{theorem}

Maintaining a fixed batch size $B_k$ as $B$, the above theorem implies that as the symplectic factor $\zeta_k$ approaches 1,  RAD converges to a point with an error bound determined by the gradient variance $\sigma^2$, i.e., $O(\sigma^2/B)$ as $N \rightarrow \infty$. Clearly, we can reduce this bound by increasing the batch size $B$. Although this does not guarantee proximity to a stationary point, in the context of machine learning, a small bound is often sufficient. To further ensure reasonable complexity, we require $\sum_{k=0}^{N-1} \frac{1}{B_{k} \sqrt{\zeta_k}}=o(N)$, which yields the following important corollary.
\begin{corollary}[Convergence of RAD]
\label{corollary_convergence}
Choosing parameters from Theorem \ref{theorem_bound} while designing $\left\{\zeta_k\right\}$ and $\left\{B_{k}\right\}$ such that $\sum_{k=0}^{N-1} \frac{1}{B_{k} \sqrt{\zeta_k}}=o(N)$, we obtain the following bound:
\begin{equation}
\nonumber
\label{eq_corollary_convergence_1}
\frac{1}{N} \sum_{k=0}^{N-1} \mathbb{E}\left\{\left\|\nabla J\left(\theta_{k}\right)\right\|^{2}\right\} \leq O\left(\frac{1}{N}\right).
\end{equation}
\end{corollary}

Corollary \ref{corollary_convergence} implies an inverse scaling relationship between the upper bound on the expected gradient norm and the number of maximum iterations, indicating the sublinear convergence of RAD. Increasing the number of maximum iterations results in a smaller upper bound and tighter optimality. Furthermore, it emphasizes that RAD's convergence can be achieved through various designs of both sequences $\left\{\zeta_k\right\}$ and $\left\{B_{k}\right\}$. In this context, we introduce a typical scheme to illustrate this concept.

\begin{corollary}
\label{corollary_design_rad_B_epsilon}
Choosing parameters from Theorem \ref{theorem_bound} while designing $\left\{\zeta_k\right\}$ and $\left\{B_{k}\right\}$ as
\begin{equation}
\label{eq_corollary_design_rad_B_epsilon}
\zeta_k = \min\{e^{{\kappa}\left(\frac{k}{{N}}-1\right)}, 1-\beta_2^{k+1}\}, \quad
B_{k}=B \times(k+1),
\end{equation}
where ${\kappa} > 0$, ${N}$ is the number of maximum iterations, and $B \in \mathbb{N}_{+}$, RAD converges to stationary points.
\end{corollary}

The configuration in Corollary \ref{corollary_design_rad_B_epsilon} allows $\zeta$ to increase slowly at the beginning, helping trainable parameters reach near-optimal space quickly. Subsequently, $\zeta$ returns to $1-\beta_2^{k+1}$, promoting symplecticity and ensuring long-term training stability. Additionally, the convergence of ADAM can be readily shown by setting $\zeta$ as a constant value. This further emphasizes the deep connection between RAD and ADAM.
\begin{corollary}[Convergence of ADAM]
\label{corollary_convergence_of_adam}
Choosing parameters from Theorem \ref{theorem_bound} while designing $\left\{\zeta_k\right\}$ and $\left\{B_{k}\right\}$ as
\begin{equation}
\label{eq_corollary_convergence_of_adam}
\zeta_k=\epsilon, \quad
B_{k}=B \times(k+1),
\end{equation}
where $\epsilon>0$ and $B\in\mathbb{N}_{+}$, RAD converges to stationary points. This exactly provides the convergence guarantee of ADAM.
\end{corollary}

In practice, we posit that a relatively modest increase in batch size is adequate to ensure the convergence of RAD. Moreover, when the gradient variance is relatively small, employing a reasonably large batch size can lead to satisfactory performance. 

\begin{remark}
Given the convergence properties of RAD in general nonconvex stochastic optimization scenarios, we anticipate extensive applications of RAD to enhance training stability across diverse machine learning domains, including but not limited to RL, where stochastic batches are utilized.
\end{remark}

\section{Numerical experiments}\label{section_experiments}
Building on our theoretical work, we conduct empirical tests on twelve benchmarks using the OpenAI Gym interface \cite{wang2023gops}. This includes one classical control task (CartPole-v1), six multi-body continuous-control tasks from MuJoCo (Ant-v3, HalfCheetah-v3, Walker2d-v3, Swimmer-v3, Hopper-v3, and Humanoid-v3) \cite{todorov2012mujoco}, four image-input discrete-control tasks from Atari games (Breakout-v4, Enduro-v4, Seaquest-v4, and SpaceInvaders-v4) \cite{bellemare2013arcade}, and one autonomous driving task using the IDSim simulator \cite{Jiang2023IDSim}. These experiments aim to validate three main hypotheses:
1) RAD stabilizes the training process of RL, thereby facilitating the attainment of high policy performance;
2) the speed coefficient $\delta$ is critical to the training stability since it controls the updating speed of parameters;
and 3) the symplectic factor $\zeta$ plays an essential role in the adaptivity of effective learning rates.

\subsection{Experimental settings}
We implement five widely used RL algorithms as testing grounds to evaluate the efficacy of RAD. These algorithms are: deep Q-network (DQN) \cite{mnih2015human}, deep deterministic policy gradient (DDPG) \cite{lillicrap2016continuous}, twin delayed DDPG (TD3) \cite{fujimoto2018addressing}, soft actor-critic (SAC) \cite{haarnoja2018soft}, and approximate dynamic programming (ADP) \cite{Powell2011ADP}. Each of these is applied to different tasks. The comparison involves six mainstream baseline optimizers: ADAM, HB, DLPF, RGD, SGD, and NAG. Additionally, to ensure relevance and currency in the rapidly evolving domains of deep learning and artificial intelligence, cutting-edge optimizers like NADAM, SWATS, and ADAMW are also included.

All experiments are run on a cloud server with one 16-core Intel$^\circledR$ Core$^\text{TM}$ i7 7820X CPU, one Nvidia$^\circledR$ Titan Xp GPU, and 62GB RAM. All experiments are conducted using an open-source RL platform, general optimal control problems solver (GOPS) \cite{wang2023gops}, which integrates RAD as a standard NN optimizer like ADAM. To ensure fairness in comparisons, we apply identical hyperparameters to all optimization algorithms involved in each task, including learning rates, momentum coefficients, batch sizes, and activation functions commonly employed in RL. Each experiment is repeated five times with a set of random seeds. The policy and the value function in Atari games are approximated using convolutional neural networks (CNNs), while multilayer perceptrons (MLPs) are employed for the other tasks. In addition, ADAM employs a default rational factor of $\epsilon=1\times10^{-16}$. Through extensive testing, we discovered that setting RAD's speed coefficient to $\delta=1$ and configuring its symplectic factor $\zeta$ to anneal according to \eqref{eq_corollary_design_rad_B_epsilon} with $\kappa=12\pi$ is effective for most tasks. This setup reduces the overhead associated with hyperparameter tuning. For additional information on training details, please consult the supplementary materials.

\subsection{Performance comparison}
\begin{figure}[!tbhp]
\centering
\subfloat[DDPG]{
\label{fig_cartpole_ddpg}
\includegraphics[width=0.45\linewidth]{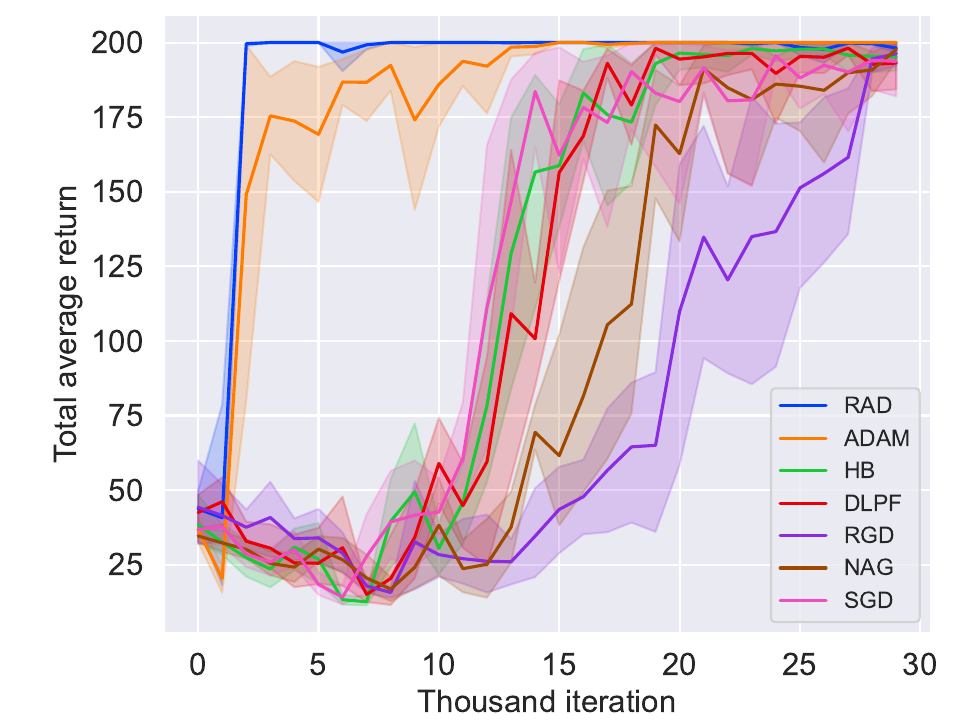}
\includegraphics[width=0.45\linewidth]{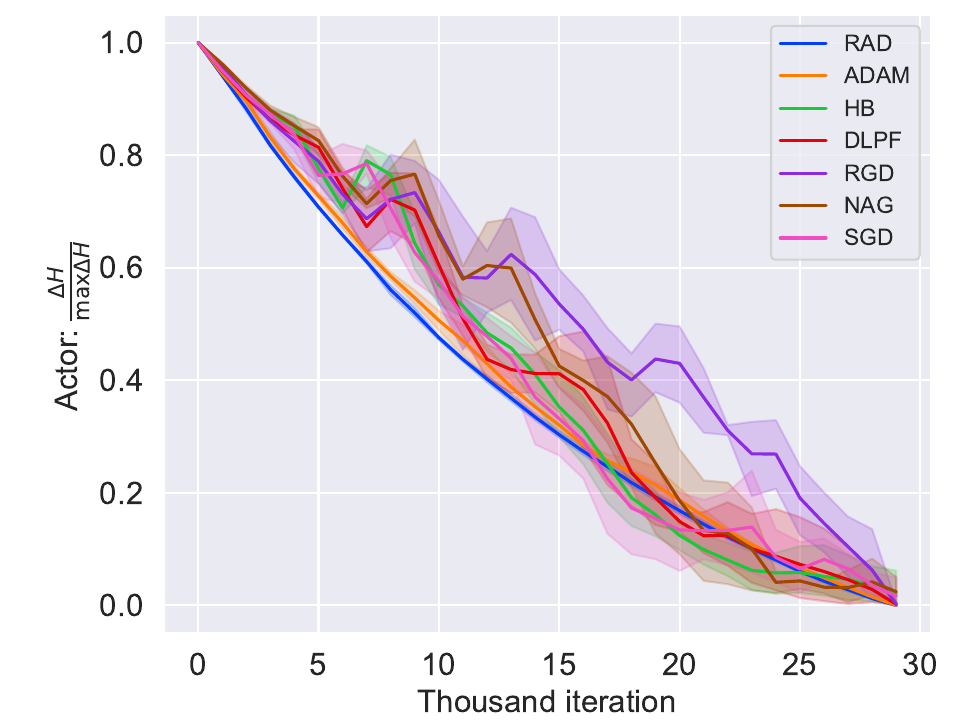}} \\
\subfloat[SAC]{
\label{fig_cartpole_sac}
\includegraphics[width=0.45\linewidth]{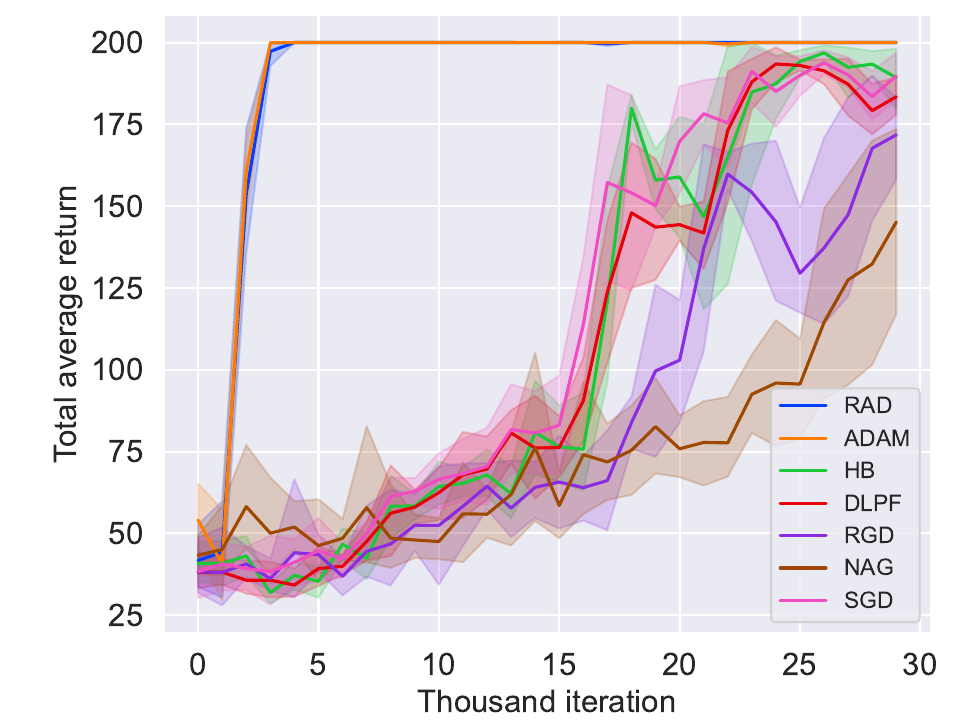}
\includegraphics[width=0.45\linewidth]{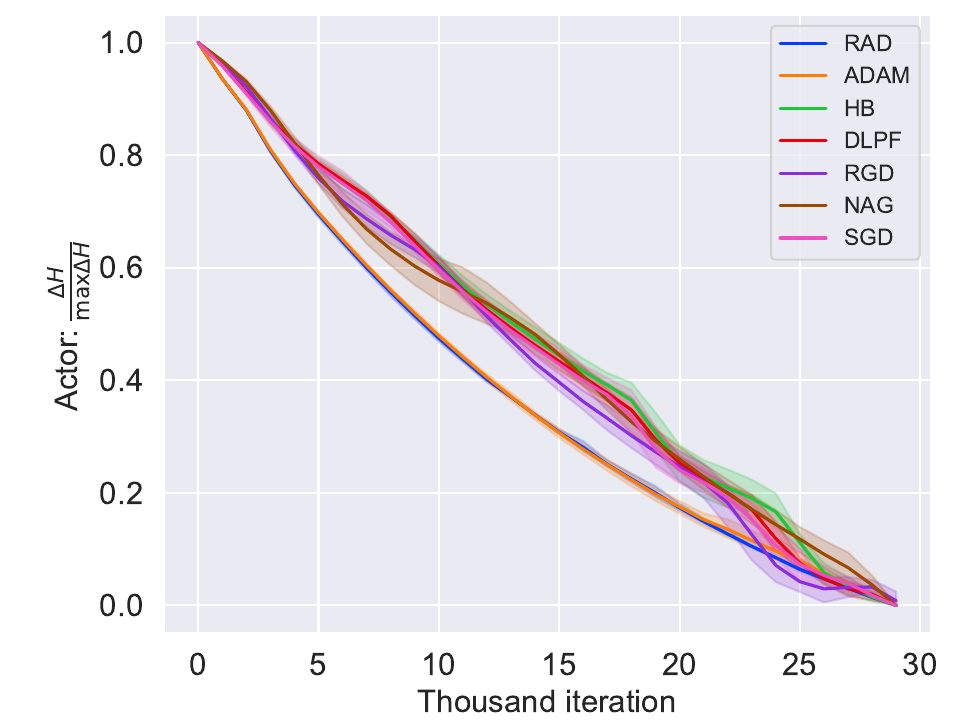}}
\caption{Learning curves of CartPole-v1, wherein the TAR indicates the policy performance and the smoothness of Hamiltonian descending implies the training stability. The solid lines correspond to the mean, the shaded regions correspond to the 95\% confidence interval over five runs, and $\Delta H=H-\min H$.}
\label{fig_cartpole task} 
\end{figure}

The learning curves of CartPole-v1 are shown in Fig.~\ref{fig_cartpole task}. The left-hand figures show the total average return (TAR) during training, i.e., the policy performance. The right-hand figures depict the decay of the Hamiltonian associated with the actor network. A monotonically and smoothly decreasing Hamiltonian signifies the dissipation of the conformal Hamiltonian system, whose dynamical properties are preserved by the corresponding conformal symplectic integrator. It is evident that RAD exhibits the smoothest curve with a monotonically descending Hamiltonian, indicating that the discrete-time dynamical system converges to stationary points with minimal oscillations. In essence, RAD effectively reproduces the dissipation of the original dynamical system while maintaining long-term stability. The TAR curve also reflects this characteristic, where the curve of RAD rises fast with barely any fluctuations. However, the curve of ADAM is more oscillating due to the lack of symplecticity. 

\begin{figure}[!thbp]
\centering
\subfloat[Ant-v3]{
\label{fig_ant_sac}
\includegraphics[width=0.45\linewidth]{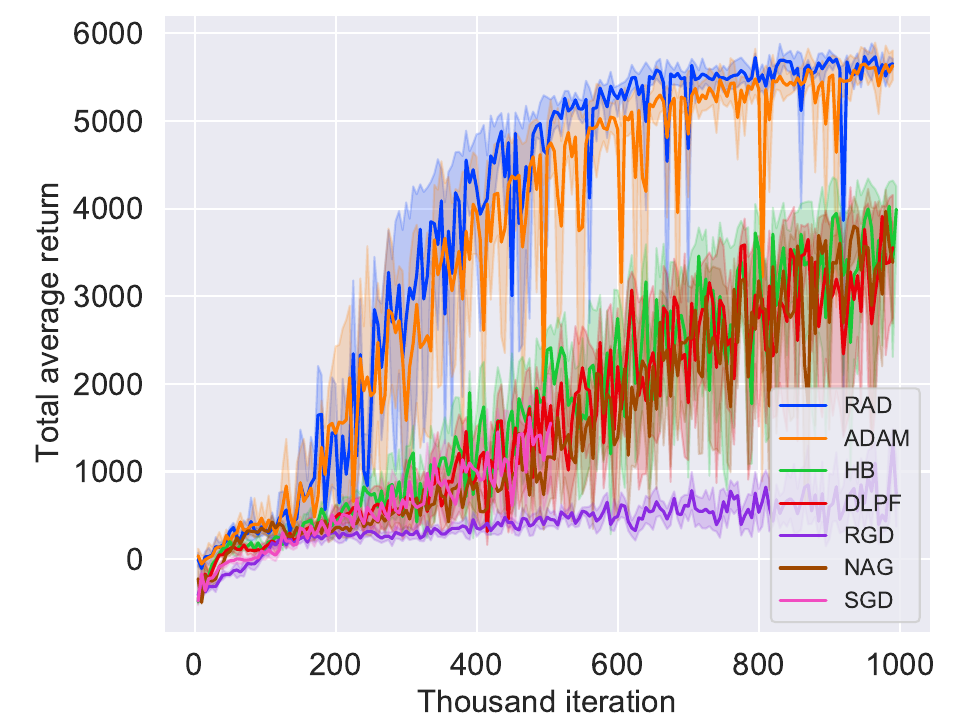}
}
\subfloat[HalfCheetah-v3]{
\label{fig_halfcheetah_sac}
\includegraphics[width=0.45\linewidth]{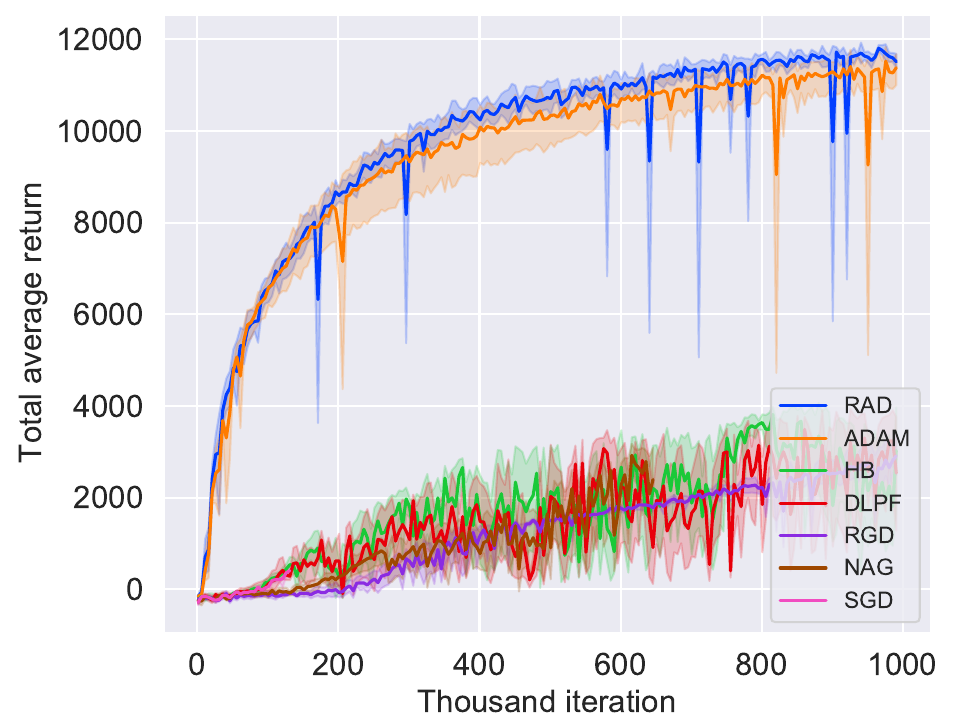}
} \\
\subfloat[Walker2d-v3]{
\label{fig_walker2d_sac}
\includegraphics[width=0.45\linewidth]{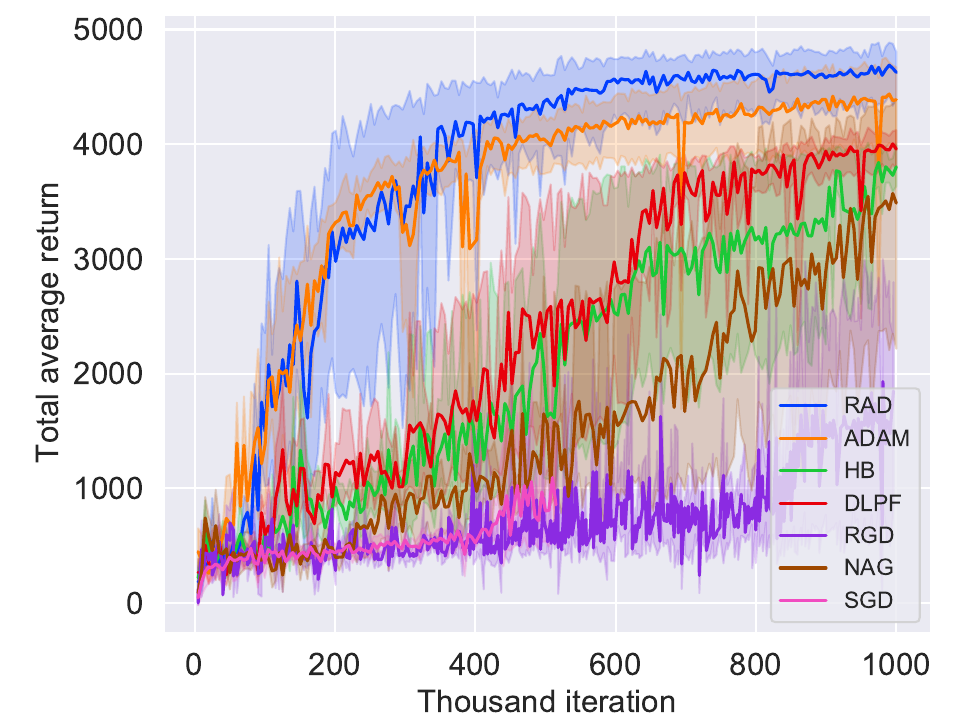}
}
\subfloat[Swimmer-v3]{
\label{fig_swimmer_sac}
\includegraphics[width=0.45\linewidth]{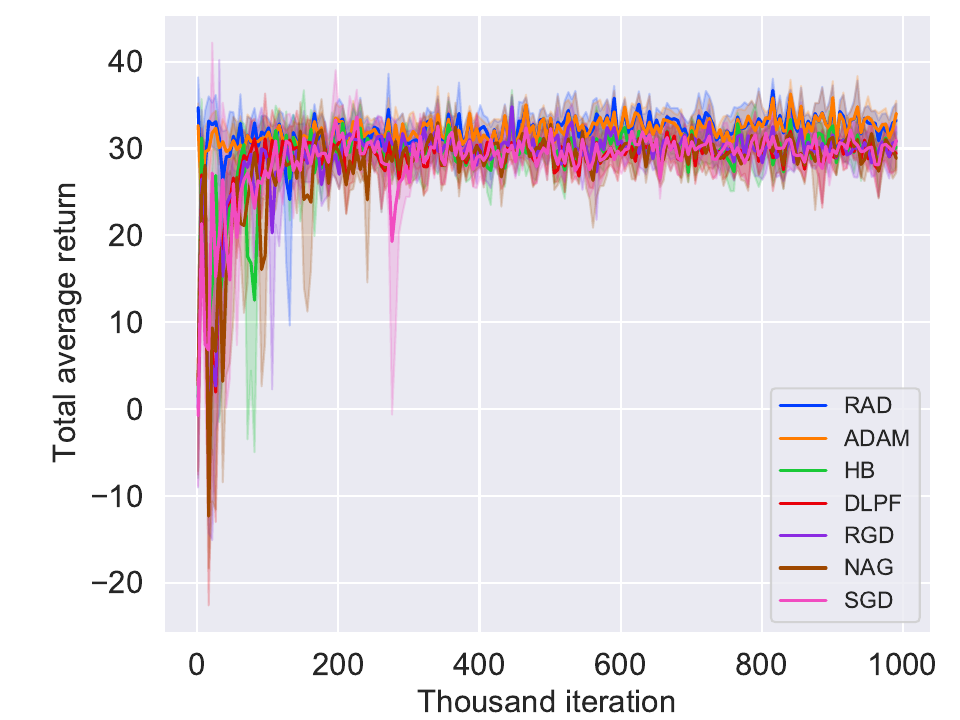}
}
\caption{
\label{fig_mujoco tasks} Policy performance of SAC on MuJoCo tasks. The solid lines correspond to the mean, and the shaded regions correspond to the 95\% confidence interval over five runs.}
\end{figure}

The policy performance on MuJoCo tasks using SAC is illustrated in Fig.~\ref{fig_mujoco tasks}. RAD and ADAM demonstrate rapid attainment of their peak TAR values, while the remaining curves exhibit a prolonged premature period before they begin to rise. RGD exhibits relatively poor performance since the effective learning rate of every parameter is adjusted with the same force, implying that it is critical to associate optimization with appropriate dynamical systems. Although NAG puts some unreal dissipation to the system aiming to accelerate training, the strong stochasticity of RL hinders its convergence. Specifically, since both are not symplectic, NAG and SGD do not preserve the long-term stability of their corresponding dynamical systems and are easy to diverge. Consequently, we can observe prematurely terminated training processes of NAG and SGD in HalfCheetah-v3 and Walker2d-v3, as evidenced by incomplete learning curves. Although the optimization processes of HB and DLPF are more stable than NAG and SGD due to their symplecticity, they converge more slowly than RAD and ADAM because of non-adaptivity. Furthermore, a stable training process is paramount for achieving optimal policy performance. Correspondingly, RAD consistently demonstrates faster convergence and achieves the highest TAR across all MuJoCo tasks, mainly exhibiting a 5.4\% improvement in TAR compared to ADAM on the Walker2d-v3 task (see Table~\ref{table_tar_sac_mujoco}). This trend is also evident in experiments conducted using TD3 (see supplementary material).

\begin{table}[!tbhp]
  \caption{Policy performance of SAC on MuJoCo tasks over five runs. The maximum value for each task is bolded. $\pm$ corresponds to the standard deviation. $\rm{DIV}$ represents early divergence. $\Uparrow$ denotes the relative improvement of RAD compared to ADAM.}
  \label{table_tar_sac_mujoco}
  \centering
  \begin{tabular}{lccccc}
    \toprule
    Tasks & Ant-v3 & HalfCheetah-v3 & Walker2d-v3 & Swimmer-v3  \\
    \midrule
    RAD & \textbf{5680}$\pm$98 & \textbf{11770}$\pm$95 & \textbf{4626}$\pm$216 & \textbf{32}$\pm$2 \\
    ADAM & 5633$\pm$204 & 11299$\pm$451 & 4388$\pm$434 & \textbf{32}$\pm$2 \\
    HB & 3993$\pm$456 & 2405$\pm$1269 & 3803$\pm$216 & 28$\pm$4 \\
    DLPF & 2669$\pm$1381 & 3310$\pm$334 & 3958$\pm$201 & 30$\pm$2 \\
    RGD & 749$\pm$488 & 2840$\pm$396 & 1634$\pm$872 & 28$\pm$3 \\
    NAG & 3634$\pm$490 & $\rm{DIV}$ & 3487$\pm$1293 & 29$\pm$2 \\
    SGD & 1932$\pm$622 & $\rm{DIV}$ & $\rm{DIV}$ & 30$\pm$2 \\
    
    $\Uparrow$ & 0.8\% & 4.2\% & 5.4\% & 0.0\% \\
    \bottomrule
  \end{tabular}
\end{table}

\begin{figure}[!htbp]
\centering
\subfloat[Breakout-v4]{
\label{fig_breakout}
\includegraphics[width=0.45\linewidth]{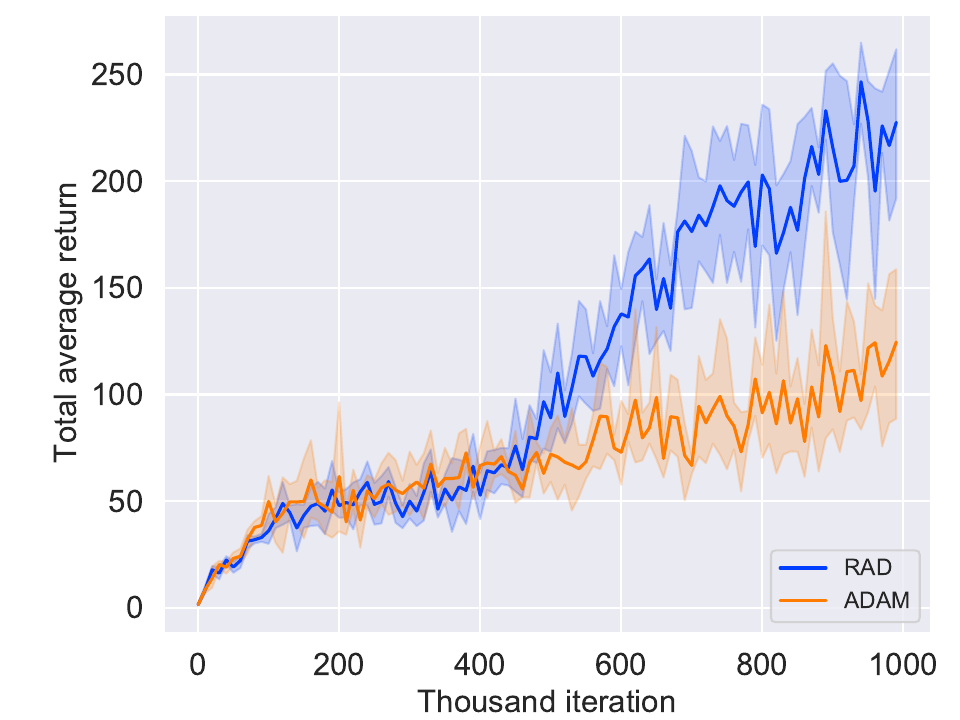}
}
\subfloat[Enduro-v4]{
\label{fig_enduro}
\includegraphics[width=0.45\linewidth]{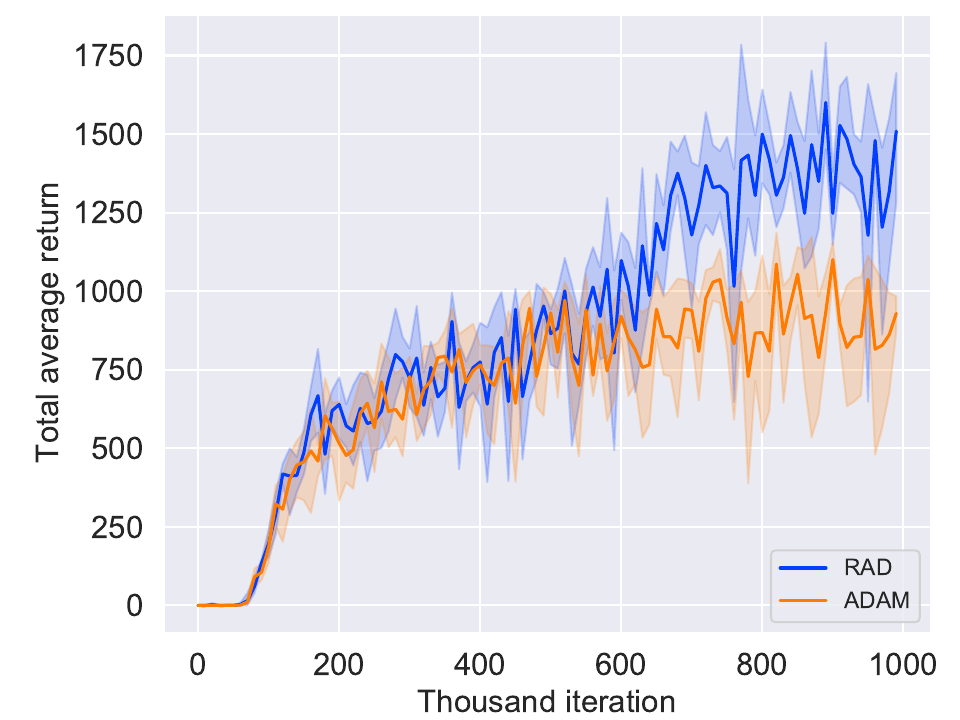}
} \\
\subfloat[Seaquest-v4]{
\label{fig_seaquest}
\includegraphics[width=0.45\linewidth]{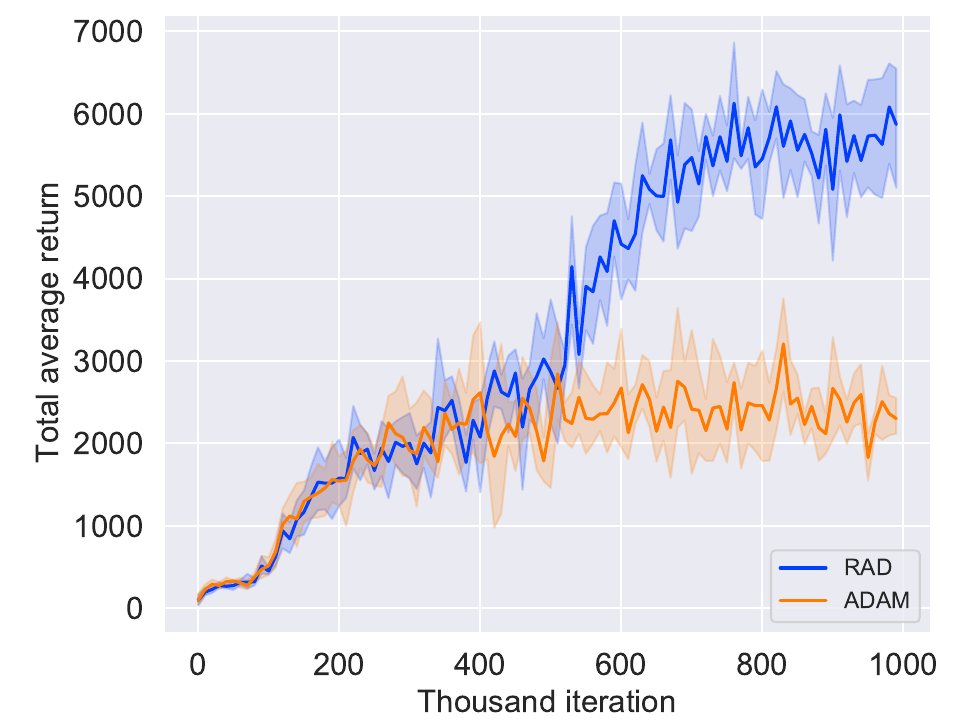}
}
\subfloat[SpaceInvaders-v4]{
\label{fig_spaceinvaders}
\includegraphics[width=0.45\linewidth]{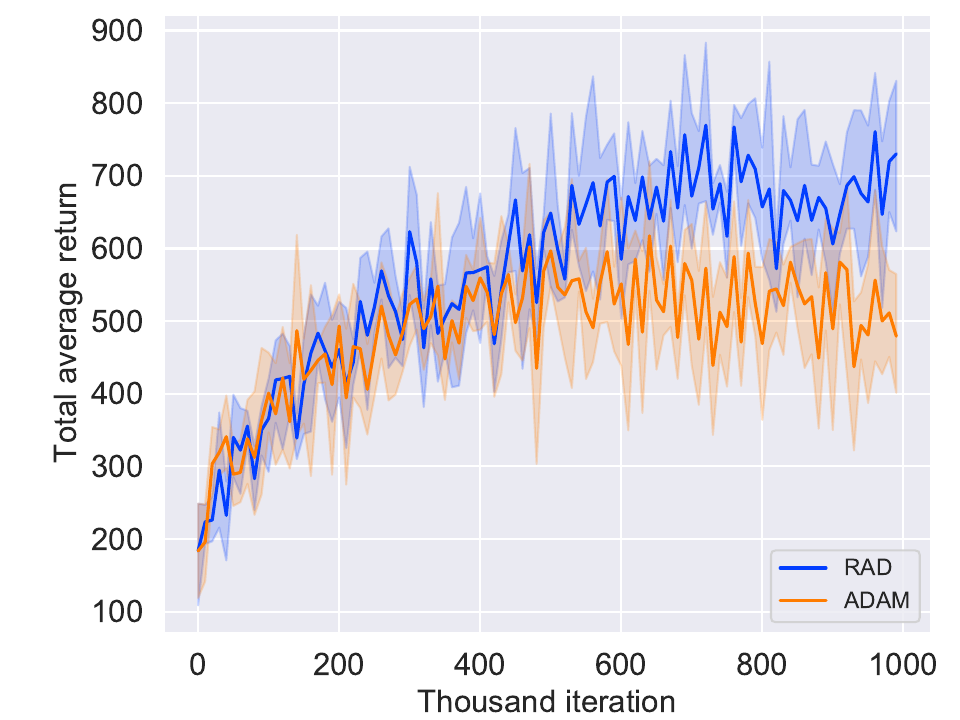}
}
\caption{
\label{fig_atari games} Policy performance on Atari games. The solid lines correspond to the mean, and the shaded regions correspond to the 95\% confidence interval over five runs.}
\end{figure}

Meanwhile, it is noteworthy that RAD shows promising potential in tasks involving image inputs. Specifically, RAD surpasses ADAM across all Atari games (refer to Fig.~\ref{fig_atari games}) and demonstrates a remarkable TAR improvement of 155.1\% on the Seaquest-v4 task (refer to Table \ref{table_tar_dqn_atari}). The effectiveness of RAD in these contexts can be attributed to its adept handling of tasks characterized by mutable gradient variance, commonly observed in image-input scenarios. Specifically, the term $y_{k+1}$ in \eqref{eq_rad_1_improved} can be likened to gradient variances. A key strength of RAD lies in its ability to mitigate the impact of mutable second-order momenta $y_{k+1}$ on the effective learning rate $\alpha_k$, achieved through a sizable symplectic factor $\zeta_k$. This mechanism stabilizes the training process, making RAD particularly robust for tasks where mutable gradient variance prevails.

\begin{table}[!tbhp]
  \caption{Policy performance of DQN on Atari games over five runs. The maximum value for each task is bolded. $\pm$ corresponds to the standard deviation. $\Uparrow$ denotes the relative improvement of RAD compared to ADAM.}
  \label{table_tar_dqn_atari}
  \centering
  \begin{tabular}{llll}
    \toprule
    Optimizer & RAD & ADAM & $\Uparrow$ \\
    \midrule
    Breakout-v4 & \textbf{228}$\pm$41 & 125$\pm$40 & 82.4\% \\
    Enduro-v4 & \textbf{1509}$\pm$245 & 929$\pm$63& 62.4\% \\
    Seaquest-v4 & \textbf{5871}$\pm$835 & 2301$\pm$247& 155.1\% \\
    SpaceInvaders-v4 & \textbf{730}$\pm$122 & 479$\pm$83& 52.4\% \\
    \bottomrule
  \end{tabular}
\end{table}

\begin{figure}[!tbhp]
\centering
\subfloat[{Walker2d-v3}]{
\label{fig_compare_walker2d}
\includegraphics[width=0.45\linewidth]{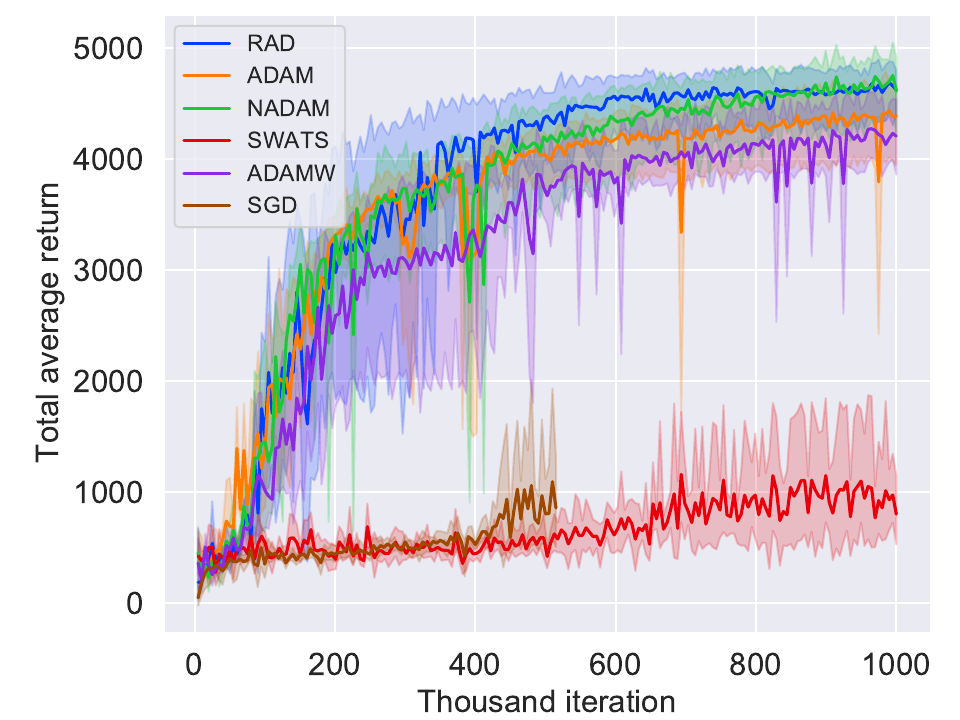}}
\subfloat[{Seaquest-v4}]{
\label{fig_compare_seaquest}
\includegraphics[width=0.45\linewidth]{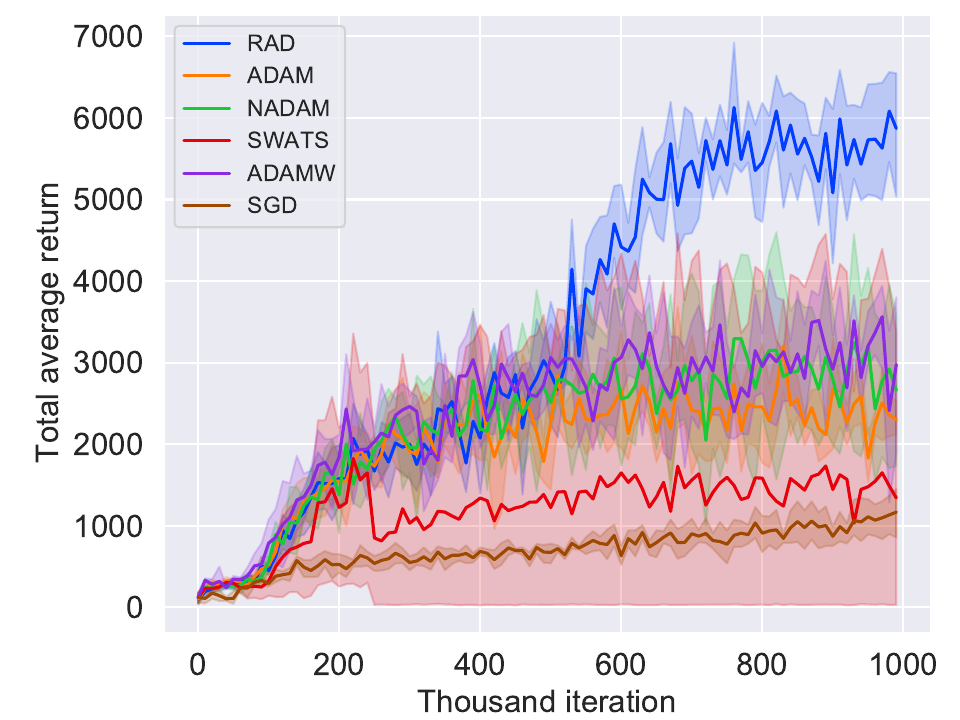}
}
\caption{\label{fig_compare_with_SOTA} {Performance comparison among SOTA optimizers. The solid lines correspond to the mean, the shaded regions correspond to the 95\% confidence interval over five runs.}}
\end{figure}

By further extending the comparison to cutting-edge optimizers, such as NADAM, SWATS, and ADAMW, it is observed that RAD consistently maintains its superiority (refer to Fig.~\ref{fig_compare_with_SOTA}). Specifically, in the Walker2d-v3 task, RAD exhibits faster convergence during the initial stage and demonstrates enhanced stability during the later stage. Moreover, RAD outperforms all competitors in the Seaquest-v4 task by a significant margin. Besides, the computational efficiency and memory usage of RAD are comparable to other adaptive methods, as they only require maintaining the momenta $v$ and $y$ during optimization. Meanwhile, it is worth noting that NADAM and ADAMW, as variants of ADAM, display relatively superior performance compared to ADAM. Since their techniques are orthometric to ours, integrating them into RAD potentially leads to further performance enhancement. We intend to explore this idea in future work.

\begin{figure}[!tbhp]
    \centering
    \includegraphics[width=0.8\linewidth]{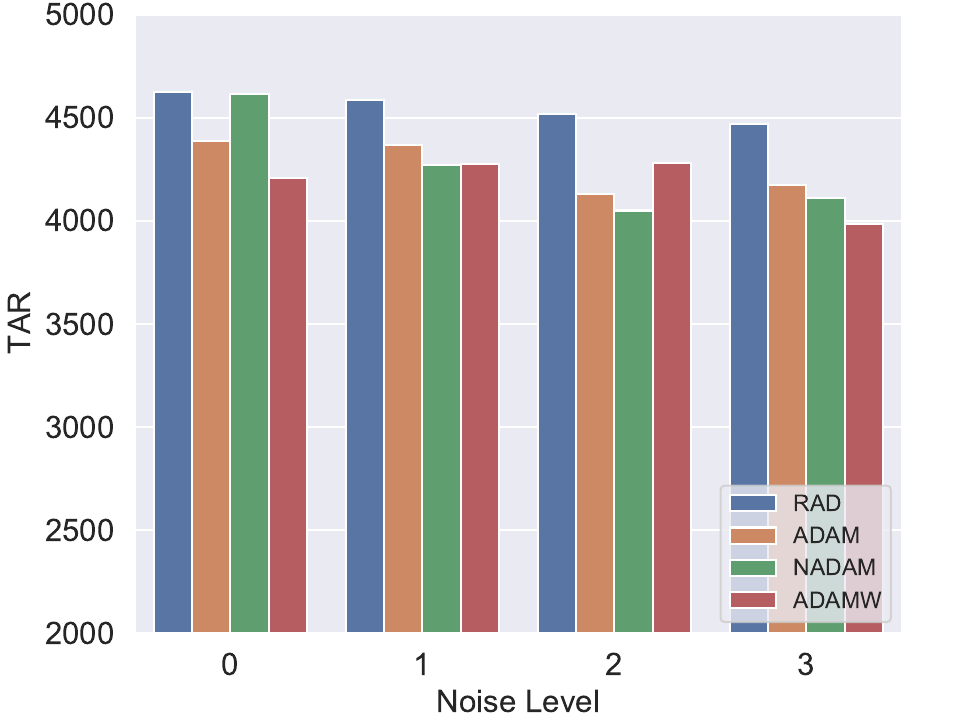}
    \caption{Policy performance on Walker2d-v3 under observation noise. For each noise level, the observation noise is randomly generated from standard Gaussian distributions with standard deviations of 0, 0.001, 0.005, and 0.01, respectively.}
    \label{fig:policy performance with noise}
\end{figure}

In addition, to evaluate the robustness of RAD, we conduct experiments on the Walker2d-v3 task with varying levels of observation noise. As shown in Fig.~\ref{fig:policy performance with noise}, RAD demonstrates a remarkable resilience to noise, thanks to the symplecticity that allows it to preserve the original conformal Hamiltonian system's evolutionary process and convergence capability. As a result, RAD exhibits a minimal performance degradation of 3.4\% under the strongest noise, in contrast to NADAM, which experiences a significantly higher degradation of 10.9\%.

Finally, to validate RAD's efficacy in handling complex and real-world RL challenges, we conduct extensive testing on demanding autonomous driving tasks. The results reveal that RAD achieves an impressive 93.2\% success rate in guiding ego vehicles through busy urban intersections, significantly outperforming ADAM's 89.5\%. This highlights RAD's adaptability and performance capabilities not only in standard RL benchmarks like MuJoCo and Atari but also in intricate real-world scenarios. Additional details on this experiment can be found in the supplementary material.

\subsection{Ablation study on different speed coefficients} 
\begin{figure}[!tbhp]
\centering
\subfloat[TAR]{
\includegraphics[width=0.45\linewidth]{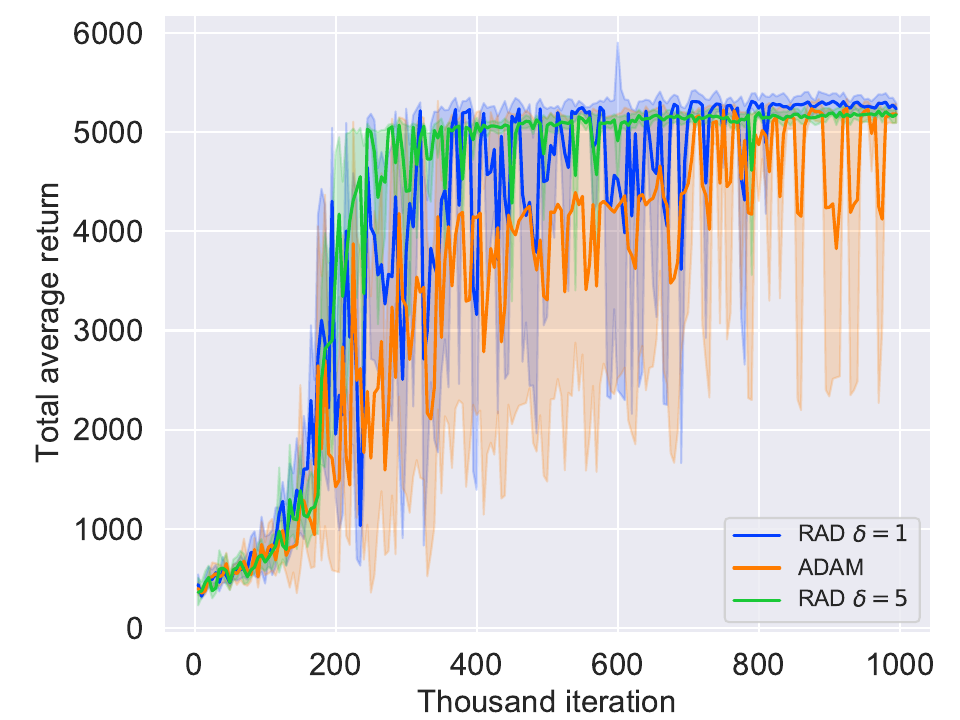}}
\subfloat[Hamiltonian]{
\includegraphics[width=0.45\linewidth]{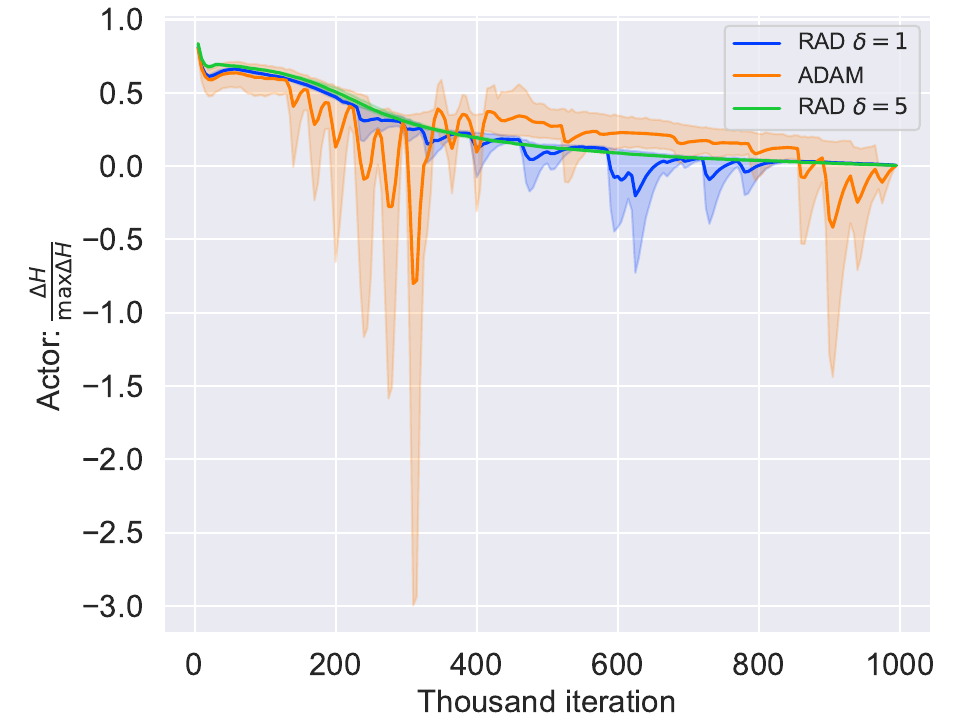}
}
\caption{\label{fig_humanoid} Ablation studies on different speed coefficients $\delta$, wherein the TAR indicates the policy performance and the smoothness of Hamiltonian descending implies the training stability. The solid lines correspond to the mean, the shaded regions correspond to the 95\% confidence interval over five runs, and $\Delta H=H-\min H$.}
\end{figure}

The Humanoid-v3 task involves a bipedal robot navigating a 376-dimensional state, aiming to walk forward without falling. This environment presents both significant complexity and large, erratic gradients. We use this task to assess RAD's stability under the influence of the speed coefficient $\delta$. As shown in Fig.~\ref{fig_humanoid}, while the TAR curve of RAD with $\delta=1$ oscillates less than ADAM, there is still potential for further improvement. Once we increase $\delta$ to 5, notable oscillation reduction is observed during the premature training stage. As discussed in Section \ref{section_rad}, $\delta$ controls the maximum speed of parameter updates and ensures that $||\theta_{k+1}-\theta_{k}||_{\infty}\leq \alpha / \delta$. A larger $\delta$ prevents rapid parameter changes, enhancing RAD's resilience against abnormal gradients. 

It is worth noting that, even with $\delta=1$, the training curve stabilizes in the mature stage due to the symplectic factor reverting to the original annealing procedure of $\zeta_k=1-\beta_2^{k+1}$, allowing RAD's symplecticity to recover. This finding suggests that the mature stage of training is not overly sensitive to $\delta$. Thus, we recommend $\delta=1$ as a default setting, as the growing symplectic factor helps mitigate potential oscillation issues. Users may still adjust $\delta$ to fine-tune performance for specific tasks or gradient behaviors.

\subsection{Ablation study on different symplectic factors} 
\begin{figure}[!tbhp]
\centering
\subfloat[TAR]{
\includegraphics[width=0.45\linewidth]{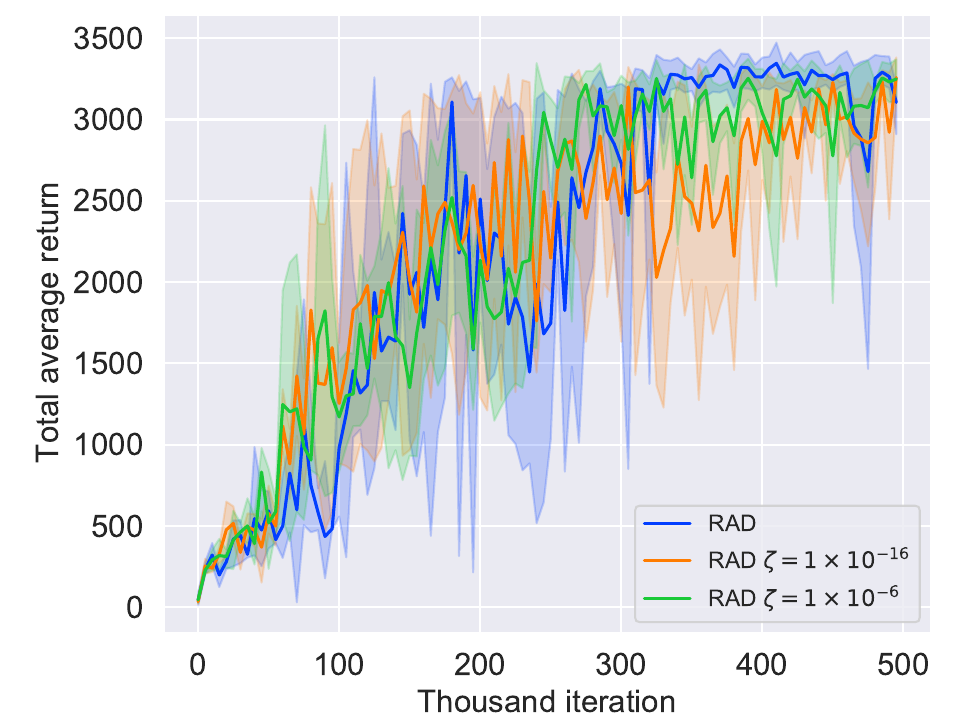}}
\subfloat[Hamiltonian]{
\includegraphics[width=0.45\linewidth]{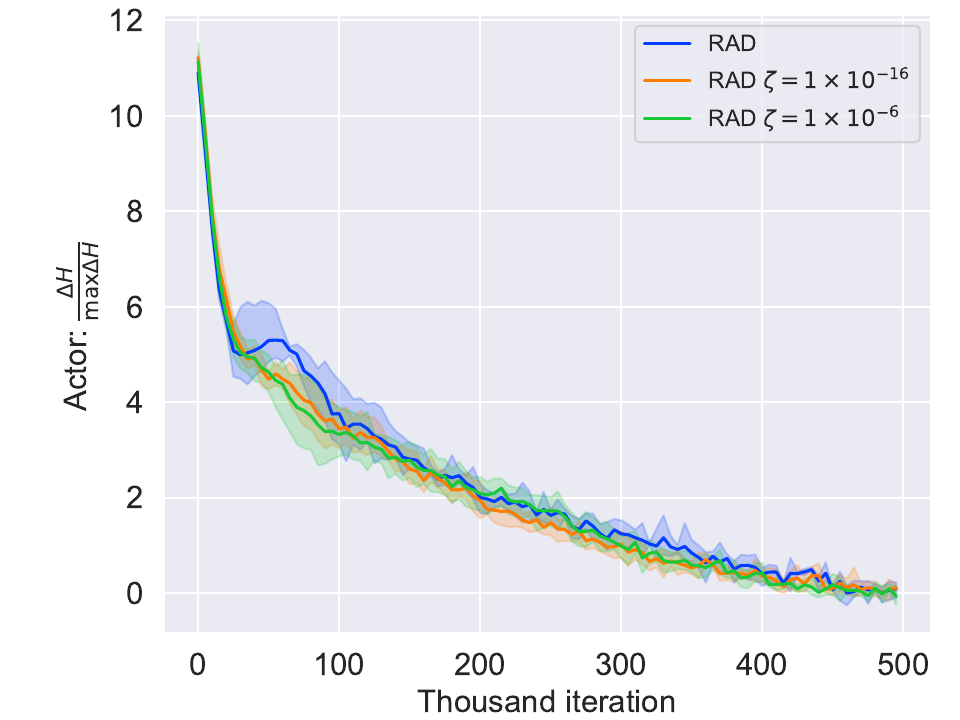}
}
\caption{\label{fig_hopper} Ablation studies on different symplectic factors $\zeta$, wherein the TAR indicates the policy performance and the smoothness of Hamiltonian descending implies the training stability. The solid lines correspond to the mean, the shaded regions correspond to the 95\% confidence interval over five runs, and $\Delta H=H-\min H$.}
\end{figure}

The Hopper-v3 task aims to make a one-legged robot hop forward as fast as possible without falling over. Abnormally large gradients with substantial variance appear in the entire training process of this task. Therefore, we choose this task as a benchmark to verify whether the symplectic factor $\zeta$ can control the adaptivity of algorithms. Only SAC is applied to this task, and the training results are shown in Fig.~\ref{fig_hopper}. Severe oscillations appear in the TAR curve of RAD with a fixed symplectic factor ${\zeta}=1\times10^{-16}$, which is precisely the case of ADAM. However, this unstable phenomenon is improved by increasing ${\zeta}$ to $1\times10^{-6}$. In fact, a reasonably sizeable ${\zeta}$ can restrain the adaptivity level of the algorithm since it reduces the influence of the second-order momenta $y_k$ on the effective learning rate $\alpha_k$, thus helping the parameter updates more robust to mutable gradients or gradients with significant variance. Therefore, with the built-in feature of an increasing symplectic factor, RAD in Algorithm \ref{section_rad} can stabilize the training process, especially at the mature stage of optimization, where $\zeta_k \rightarrow 1-\beta_2^{k+1}$ enhances the symplecticity of RAD.

\section{Conclusion}\label{section_conclusion}
This paper systematically investigates the optimization process of training an RL agent from a dynamic systems perspective. To improve the stability of RL training, we model NN optimization as the evolution of conformal Hamiltonian systems. By discretizing the system in a conformally symplectic manner, we develop iterative updating rules that incorporate symplecticity and long-term stability into the optimization process. Our primary contribution is the RAD algorithm, which simulates a relativistic system governed by multiple one-dimensional particles. RAD adapts and limits the updating for each trainable parameter individually, promoting stable and rapid convergence. Our findings demonstrate that RAD generalizes the widely used ADAM algorithm, providing insights into the inherent dynamics of other mainstream adaptive gradient algorithms. We establish RAD's convergence properties under general nonconvex stochastic optimization settings to ensure broader applicability. Experimental results confirm that RAD achieves state-of-the-art performance under default settings, highlighting its potential to stabilize RL training. Future work may explore the shadow dynamics of current mainstream methods to deepen understanding and enhance RAD's performance by integrating techniques from cutting-edge optimizers.

\clearpage

\appendix
\subsection{Proof of Corollary \ref{corollary_3}}
Under Assumption \ref{assumption_1} and Assumption \ref{assumption_2}, if given any $\theta_1 \in \mathbb{R}^{n}$ and $\theta_2 \in \mathbb{R}^{n}$, we have
\begin{displaymath}
\begin{aligned}
&\left\|\nabla J\left(\theta_{1}\right)-\nabla J\left(\theta_{2}\right)\right\| \\
=& \left\|\nabla \mathbb{E}_{x \sim \mathcal{P}}\left\{\mathcal{L}\left(x, \theta_{1}\right)\right\}-\nabla \mathbb{E}_{x \sim \mathcal{P}}\left\{\mathcal{L}\left(x, \theta_{2}\right)\right\}\right\| \\
=& \left\|\mathbb{E}_{x \sim \mathcal{P}}\left\{\nabla \mathcal{L}\left(x, \theta_{1}\right)\right\}-\mathbb{E}_{x \sim \mathcal{P}}\left\{\nabla \mathcal{L}\left(x, \theta_{2}\right)\right\}\right\| \\
=& \left\|\mathbb{E}_{x \sim \mathcal{P}}\left\{\nabla \mathcal{L}\left(x, \theta_{1}\right)-\nabla \mathcal{L}\left(x, \theta_{2}\right)\right\}\right\| \\
\leq&  \mathbb{E}_{x \sim \mathcal{P}}\left\{\left\|\nabla \mathcal{L}\left(x, \theta_{1}\right)-\nabla \mathcal{L}\left(x, \theta_{2}\right)\right\|\right\} \\
\leq&  \mathbb{E}_{x \sim \mathcal{P}}\left\{L\left\|\theta_{1}-\theta_{2}\right\|\right\} \\
=& L\left\|\theta_{1}-\theta_{2}\right\|.
\end{aligned}
\end{displaymath}

\subsection{Proof of Corollary \ref{corollary_4}}
Under Assumption \ref{assumption_1} and Assumption \ref{assumption_2}, if given any $\theta \in \mathbb{R}^{n}$ and $i \in \{1,2,\cdots,n\}$, we have
\begin{displaymath}
\begin{aligned}
\lvert\left[\nabla J(\theta)\right]_{i}\rvert &=\lvert\mathbb{E}_{x \sim \mathcal{P}}\{[\nabla \mathcal{L}(x, \theta)]_i\}\rvert \\
&\leq \mathbb{E}_{x \sim \mathcal{P}}\{\lvert[\nabla \mathcal{L}(x, \theta)]_i\rvert\} \leq M.
\end{aligned}
\end{displaymath}

\subsection{Proof of Corollary \ref{corollary_5}}
Under Assumption \ref{assumption_1}, Assumption \ref{assumption_2}, and Assumption \ref{assumption_3}, if given any $\theta \in \mathbb{R}^{n}$, we have
\begin{displaymath}
\begin{aligned}
&\mathbb{E}_{x \sim \mathcal{P}}\left\{\|\nabla \mathcal{L}(x, \theta)-\nabla J(\theta)\|^{2}\right\} \\
=& \mathbb{E}_{x \sim \mathcal{P}}\left\{\sum_{i=1}^{n}\left([\nabla \mathcal{L}(x, \theta)]_{i}-[\nabla J(\theta)]_{i}\right)^{2}\right\} \\
=& \sum_{i=1}^{n} \mathbb{E}_{x \sim \mathcal{P}}\left\{\left([\nabla \mathcal{L}(x, \theta)]_{i}-[\nabla J(\theta)]_{i}\right)^{2}\right\} \\
\leq&  \sum_{i=1}^{n} \sigma_{i}^{2} =\sigma^{2}.
\end{aligned}
\end{displaymath}

\subsection{Proof of Corollary \ref{corollary_convergence}}
By substituting $\sum_{k=0}^{N-1} \frac{1}{B_{k} \sqrt{\zeta_k}}=o(N)$ into \eqref{eq_theorem_bound}, we immediately obtain the following bound of RAD:
\begin{equation}
\nonumber
\frac{1}{N} \sum_{k=0}^{N-1} \mathbb{E}\left\{\left\|\nabla J\left(\theta_{k}\right)\right\|^{2}\right\} \leq O\left(\frac{1}{N}\right).
\end{equation}

\subsection{Proof of Corollary \ref{corollary_design_rad_B_epsilon}}
Let us assume that $\zeta_k$ switches back to $1-\beta_2^{k+1}$ at the $\hat{N}$-th iteration. According to Corollary \ref{corollary_convergence}, if substituting \eqref{eq_corollary_design_rad_B_epsilon} into $\sum_{k=0}^{N-1} \frac{1}{B_{k} \sqrt{\zeta_k}}$, we have 
\begin{displaymath}
\begin{aligned}
\sum_{k=0}^{N-1} \frac{1}{B_{k} \sqrt{\zeta_k}} & =\frac{1}{B} \sum_{k=0}^{N-1} \frac{1}{(k+1) \sqrt{\zeta_k}} \\
& =\frac{1}{B}\left(\sum_{k=0}^{\hat{N}-1} \frac{1}{(k+1) \sqrt{\zeta_k}}+\sum_{k=\hat{N}}^{N-1} \frac{1}{(k+1) \sqrt{\zeta_k}}\right) \\
& \leq \frac{1}{B}\left(\sum_{k=0}^{\hat{N}-1} \frac{1}{\sqrt{\zeta_k}}+\sum_{k=\hat{N}}^{N-1} \frac{1}{k+1}\right) \\
& = \frac{1}{B}\left(\sum_{k=0}^{\hat{N}-1} e^{\frac{\kappa}{2}\left(1-\frac{k}{\hat{N}}\right)}+\sum_{k=\hat{N}}^{N-1} \frac{1}{k+1}\right) \\
& \leq \frac{1}{B}\left(\sum_{k=0}^{\hat{N}-1} e^{\frac{\kappa}{2}\left(1-\frac{k}{\hat{N}}\right)}+\sum_{k=\hat{N}}^{N-1} \ln \frac{k+1}{k}\right) \\
& = \frac{1}{B}\left(\sum_{k=0}^{\hat{N}-1} e^{\frac{\kappa}{2}\left(1-\frac{k}{\hat{N}}\right)}+ \ln{N} - \ln{\hat{N}}\right) \\
&=\frac{1}{B} \ln N-\frac{1}{B}\left(\ln \hat{N}-\sum_{k=0}^{\hat{N}-1} e^{\frac{\kappa}{2}\left(1-\frac{k}{\hat{N}}\right)}\right) \\
&= o(N).
\end{aligned}
\end{displaymath}

\subsection{Proof of Corollary \ref{corollary_convergence_of_adam}}
According to Corollary \ref{corollary_convergence}, if substituting \eqref{eq_corollary_convergence_of_adam} into $\sum_{k=0}^{N-1} \frac{1}{B_{k} \sqrt{\zeta_k}}$, we have
\begin{displaymath}
\begin{aligned}
\sum_{k=0}^{N-1} \frac{1}{B_{k}\sqrt{\epsilon}} =&\frac{1}{B \sqrt{\epsilon}} \sum_{k=0}^{N-1} \frac{1}{k+1} \\
\leq& \frac{1}{B \sqrt{\epsilon}}(\ln N+C+\varepsilon) \\
=& o(N),
\end{aligned}
\end{displaymath}
where $C$ is the Euler's constant and $\varepsilon>0$ is an arbitrary positive number. 

\subsection{Proof of Lemma \ref{lemma_1}}
We define the $i$-th coordinate of the stochastic gradient $g_k$ calculated at the $k$-th iteration by using the mini-batch $\mathcal{B}_k$ of which size is $B_k$ as
\begin{displaymath}
g_{k, i}=\frac{1}{B_{k}} \sum_{x \in \mathcal{B}_{k}}\left[\nabla \mathcal{L}\left(x, \theta_{k}\right)\right]_{i}.
\end{displaymath}

Let us further define the following notation for ease of explanation:
\begin{displaymath}
\xi_{k, i}=\frac{1}{B_{k}} \sum_{x \in \mathcal{B}_{k}}\left(\left[\nabla \mathcal{L}\left(x, \theta_{k}\right)\right]_{i}-\left[\nabla J\left(\theta_{k}\right)\right]_{i}\right) .
\end{displaymath}

Using this notation, we have
\begin{displaymath}
\begin{aligned}
&\mathbb{E}\left\{g_{k, i}^{2} \mid \theta_{k}\right\}\\
=& \mathbb{E}\left\{\left(\xi_{k, i}+\left[\nabla J\left(\theta_{k}\right)\right]_{i}\right)^{2} \mid \theta_{k}\right\} \\
=& \mathbb{E}\left\{\xi_{k, i}^{2} \mid \theta_{k}\right\}+2\left[\nabla J\left(\theta_{k}\right)\right]_{i} \mathbb{E}\left\{\xi_{k, i} \mid \theta_{k}\right\}+\left[\nabla J\left(\theta_{k}\right)\right]_{i}^{2} \\
=& \mathbb{E}\left\{\xi_{k, i}^{2} \mid \theta_{k}\right\}+\left[\nabla J\left(\theta_{k}\right)\right]_{i}^{2} \\
=& \frac{1}{B_{k}^{2}} \mathbb{E}\left\{\left(\sum_{x \in \mathcal{B}_{k}}\left(\left[\nabla \mathcal{L}\left(x, \theta_{k}\right)\right]_{i}-\left[\nabla J\left(\theta_{k}\right)\right]_{i}\right)\right)^{2} \mid \theta_{k}\right\}+\left[\nabla J\left(\theta_{k}\right)\right]_{i}^{2} \\
=& \frac{1}{B_{k}^{2}} \mathbb{E}\left\{\sum_{x \in \mathcal{B}_{k}}\left(\left[\nabla \mathcal{L}\left(x, \theta_{k}\right)\right]_{i}-\left[\nabla J\left(\theta_{k}\right)\right]_{i}\right)^{2} \mid \theta_{k}\right\}+\left[\nabla J\left(\theta_{k}\right)\right]_{i}^{2} \\
=& \frac{1}{B_{k}^{2}} \sum_{x \in \mathcal{B}_{k}} \mathbb{E}\left\{\left(\left[\nabla \mathcal{L}\left(x, \theta_{k}\right)\right]_{i}-\left[\nabla J\left(\theta_{k}\right)\right]_{i}\right)^{2} \mid \theta_{k}\right\}+\left[\nabla J\left(\theta_{k}\right)\right]_{i}^{2} \\
\leq&  \frac{1}{B_{k}} \sigma_{i}^{2}+\left[\nabla J\left(\theta_{k}\right)\right]_{i}^{2}.
\end{aligned}
\end{displaymath}
The third equation is due to the fact that $\xi_{k, i}$ is a zero-mean random variable. The fifth equation is due to the independence among $\left\{\left[\nabla \mathcal{L}\left(\mathcal{B}_{k, j}, \theta_{k}\right)\right]_{i}-\left[\nabla J\left(\theta_{k}\right)\right]_{i}\right\}$ and the fact that $\mathbb{E}\left\{\left[\nabla \mathcal{L}\left(\mathcal{B}_{k, j}, \theta_{k}\right)\right]_{i}-\left[\nabla J\left(\theta_{k}\right)\right]_{i} \mid \theta_k \right\}=0$ for any $\theta_k$ and $j\in\{1,2, \cdots, B_{k}\}$.

\subsection{Proof of Theorem \ref{theorem_bound}}
Recall that the element-wise parameter updating rule of the first-order RAD (Algorithm \ref{algorithm_rad_final}) is
\begin{displaymath}
\theta_{k+1, i}=\theta_{k, i}-\alpha \frac{g_{k, i}}{\sqrt{\delta^{2} y_{k+1, i}+\zeta_k}}, i \in \{1,2, \cdots, n\}.
\end{displaymath}

Here $g_{k,i}$ is the $i$-th coordinate of stochastic gradient $g_k$ by computing the batch $\mathcal{B}_k$ with size $B_k$ at the $k$-th iteration. According to Corollary \ref{corollary_3}, the objective function $J$ is L-smooth, we have
\begin{displaymath}
\begin{aligned}
J\left(\theta_{k+1}\right) \leq & J\left(\theta_{k}\right)+\nabla^\top J\left(\theta_{k}\right)\left(\theta_{k+1}-\theta_{k}\right)+\frac{L}{2}\left\|\theta_{k+1}-\theta_{k}\right\|^{2} \\
=& J\left(\theta_{k}\right)-\alpha \sum_{i=1}^{n}\left(\left[\nabla J\left(\theta_{k}\right)\right]_{i} \times \frac{g_{k, i}}{\sqrt{\delta^{2} y_{k+1, i}+\zeta_k}}\right)\\
&+\frac{L \alpha^{2}}{2} \sum_{i=1}^{n} \frac{g_{k, i}^{2}}{\delta^{2} y_{k+1, i}+\zeta_k}.
\end{aligned}
\end{displaymath}

Given $\theta_k$, we take the conditional expectation with respect to the batch $\mathcal{B}_k$ on both sides of the above inequation, thus having
\begin{displaymath}
\begin{aligned}
&\mathbb{E}\left\{J\left(\theta_{k+1}\right) \mid \theta_{k}\right\} \\
\leq &J\left(\theta_{k}\right)-\alpha \sum_{i=1}^{n}\left(\left[\nabla J\left(\theta_{k}\right)\right]_{i} \times \mathbb{E}\left\{\frac{g_{k, i}}{\sqrt{\delta^{2} y_{k+1, i}+\zeta_k}} \mid \theta_{k}\right\}\right) \\
&+\frac{L \alpha^{2}}{2} \sum_{i=1}^{n} \mathbb{E}\left\{\frac{g_{k, i}^{2}}{\delta^{2} y_{k+1, i}+\zeta_k} \mid \theta_{k}\right\}\\
\end{aligned}
\end{displaymath}
\begin{displaymath}
\begin{aligned}
=&J\left(\theta_{k}\right)-\alpha \sum_{i=1}^{n}\left(\left[\nabla J\left(\theta_{k}\right)\right]_{i} \times \mathbb{E}\left\{\frac{g_{k, i}}{\sqrt{\delta^{2} \beta_{2} y_{k, i}+\zeta_k}} \right.\right.\\
&\left.\left.+\frac{g_{k, i}}{\sqrt{\delta^{2} y_{k+1, i}+\zeta_k}}-\frac{g_{k, i}}{\sqrt{\delta^{2} \beta_{2} y_{k, i}+\zeta_k}} \mid \theta_{k}\right\}\right)\\
&+\frac{L \alpha^{2}}{2} \sum_{i=1}^{n} \mathbb{E}\left\{\frac{g_{k, i}^{2}}{\delta^{2} y_{k+1, i}+\zeta_k} \mid \theta_{k}\right\}\\
=&J\left(\theta_{k}\right)-\alpha \sum_{i=1}^{n}\left(\left[\nabla J\left(\theta_{k}\right)\right]_{i} \times\left(\frac{\left[\nabla J\left(\theta_{k}\right)\right]_{i}}{\sqrt{\delta^{2} \beta_{2} y_{k, i}+\zeta_k}} \right.\right.\\
&\left.\left.+\mathbb{E}\left\{\frac{g_{k, i}}{\sqrt{\delta^{2} y_{k+1, i}+\zeta_k}}-\frac{g_{k, i}}{\sqrt{\delta^{2} \beta_{2} y_{k, i}+\zeta_k}} \mid \theta_{k}\right\}\right)\right)\\
&+\frac{L \alpha^{2}}{2} \sum_{i=1}^{n} \mathbb{E}\left\{\frac{g_{k, i}^{2}}{\delta^{2} y_{k+1, i}+\zeta_k} \mid \theta_{k}\right\}\\
=&J\left(\theta_{k}\right)-\alpha \sum_{i=1}^{n} \frac{\left[\nabla J\left(\theta_{k}\right)\right]_{i}^{2}}{\sqrt{\delta^{2} \beta_{2} y_{k, i}+\zeta_k}} -\alpha \sum_{i=1}^{n}\left(\left[\nabla J\left(\theta_{k}\right)\right]_{i} \right.\\
&\left. \times \mathbb{E}\left\{\frac{g_{k, i}}{\sqrt{\delta^{2} y_{k+1, i}+\zeta_k}}-\frac{g_{k, i}}{\sqrt{\delta^{2} \beta_{2} y_{k, i}+\zeta_k}} \mid \theta_{k}\right\}\right)\\
&+\frac{L \alpha^{2}}{2} \sum_{i=1}^{n} \mathbb{E}\left\{\frac{g_{k, i}^{2}}{\delta^{2} y_{k+1, i}+\zeta_k} \mid \theta_{k}\right\}\\
\leq& J\left(\theta_{k}\right)-\alpha \sum_{i=1}^{n} \frac{\left[\nabla J\left(\theta_{k}\right)\right]_{i}^{2}}{\sqrt{\delta^{2} \beta_{2} y_{k, i}+\zeta_k}} +\alpha \sum_{i=1}^{n}\left(\lvert\left[\nabla J\left(\theta_{k}\right)\right]_{i}\rvert \right.\\
&\left.\times\lvert\mathbb{E}\left\{\frac{g_{k, i}}{\sqrt{\delta^{2} y_{k+1, i}+\zeta_k}}-\frac{g_{k, i}}{\sqrt{\delta^{2} \beta_{2} y_{k, i}+\zeta_k}} \mid \theta_{k}\right\}\rvert\right)\\
&+\frac{L \alpha^{2}}{2} \sum_{i=1}^{n} \mathbb{E}\left\{\frac{g_{k, i}^{2}}{\delta^{2} y_{k+1, i}+\zeta_k} \mid \theta_{k}\right\} .
\end{aligned}
\end{displaymath}

The second equation follows from the fact that $g_k$ is an unbiased estimate of $\nabla J\left(\theta_k\right)$, i.e., $\mathbb{E}\left\{g_{k}\right\}=\nabla J\left(\theta_{k}\right)$, and $y_k$ is independent of $\mathcal{B}_k$ sampled at the $k$-th step. We define
\begin{displaymath}
T_{1}=\frac{g_{k, i}}{\sqrt{\delta^{2} y_{k+1, i}+\zeta_k}}-\frac{g_{k, i}}{\sqrt{\delta^{2} \beta_{2} y_{k, i}+\zeta_k}}.
\end{displaymath}

Then, an upper bound of $T_1$ is
\begin{displaymath}
\begin{aligned}
T_{1} =&\frac{g_{k, i}}{\sqrt{\delta^{2} y_{k+1, i}+\zeta_k}}-\frac{g_{k, i}}{\sqrt{\delta^{2} \beta_{2} y_{k, i}+\zeta_k}} \\
\leq&\lvert g_{k, i}\rvert \times\lvert\frac{1}{\sqrt{\delta^{2} y_{k+1, i}+\zeta_k}}-\frac{1}{\sqrt{\delta^{2} \beta_{2} y_{k, i}+\zeta_k}}\rvert \\
=&\frac{\lvert g_{k, i}\rvert}{\sqrt{\left(\delta^{2} y_{k+1, i}+\zeta_k\right)\left(\delta^{2} \beta_{2} y_{k, i}+\zeta_k\right)}} \\
&\times\lvert\sqrt{\delta^{2} \beta_{2} y_{k, i}+\zeta_k}-\sqrt{\delta^{2} y_{k+1, i}+\zeta_k}\rvert \\
=&\frac{\lvert g_{k, i}\rvert}{\sqrt{\left(\delta^{2} y_{k+1, i}+\zeta_k\right)\left(\delta^{2} \beta_{2} y_{k, i}+\zeta_k\right)}} \\
&\times\lvert\frac{\delta^{2}\left(y_{k+1, i}-\beta_{2} y_{k, i}\right)}{\sqrt{\delta^{2} y_{k+1, i}+\zeta_k}+\sqrt{\delta^{2} \beta_{2} y_{k, i}+\zeta_k}}\rvert \\
=&\frac{\lvert g_{k, i}\rvert}{\sqrt{\left(\delta^{2} y_{k+1, i}+\zeta_k\right)\left(\delta^{2} \beta_{2} y_{k, i}+\zeta_k\right)}} \\
&\times \frac{\delta^{2}\left(1-\beta_{2}\right) g_{k, i}^{2}}{\sqrt{\delta^{2} y_{k+1, i}+\zeta_k}+\sqrt{\delta^{2} \beta_{2} y_{k, i}+\zeta_k}}.
\end{aligned}
\end{displaymath}

The fourth equation is due to the updating rule of $y$ in RAD, i.e., $y_{k+1}=\beta_{2} y_{k}+\left(1-\beta_{2}\right) g_{k}^{2}$. We further bound $T_1$ as follows:
\begin{displaymath}
\begin{aligned}
T_{1} \leq &\frac{\lvert g_{k, i}\rvert}{\sqrt{\left(\delta^{2} y_{k+1, i}+\zeta_k\right)\left(\delta^{2} \beta_{2} y_{k, i}+\zeta_k\right)}} \\
&\times \frac{\delta^{2}\left(1-\beta_{2}\right) g_{k, i}^{2}}{\sqrt{\delta^{2} y_{k+1, i}+\zeta_k}+\sqrt{\delta^{2} \beta_{2} y_{k, i}+\zeta_k}}\\
=&\frac{\lvert g_{k, i}\rvert}{\sqrt{\delta^{2} y_{k+1, i}\left(\delta^{2} \beta_{2} y_{k, i}+\zeta_k\right)+\zeta_k\left(\delta^{2} \beta_{2} y_{k, i}+\zeta_k\right)}} \\
&\times \frac{\delta^{2}\left(1-\beta_{2}\right) g_{k, i}^{2}}{\sqrt{\delta^{2}\left(\beta_{2} y_{k, i}+\left(1-\beta_{2}\right) g_{k, i}^{2}\right)+\zeta_k}+\sqrt{\delta^{2} \beta_{2} y_{k, i}+\zeta_k}}\\
\leq &\frac{\lvert g_{k, i}\rvert}{\sqrt{\zeta_k\left(\delta^{2} \beta_{2} y_{k, i}+\zeta_k\right)}} \times \frac{\delta^{2}\left(1-\beta_{2}\right) g_{k, i}^{2}}{\sqrt{\delta^{2}\left(1-\beta_{2}\right) g_{k, i}^{2}}}\\
=&\frac{\delta \sqrt{1-\beta_{2}} g_{k, i}^{2}}{\sqrt{\zeta_k\left(\delta^{2} \beta_{2} y_{k, i}+\zeta_k\right)}}.
\end{aligned}
\end{displaymath}

Therefore, we have
\begin{displaymath}
\begin{aligned}
&\mathbb{E}\left\{J\left(\theta_{k+1}\right) \mid \theta_{k}\right\} \\
\leq &J\left(\theta_{k}\right)-\alpha \sum_{i=1}^{n} \frac{\left[\nabla J\left(\theta_{k}\right)\right]_{i}^{2}}{\sqrt{\delta^{2} \beta_{2} y_{k, i}+\zeta_k}} \\
&+\alpha \sum_{i=1}^{n}\left(\lvert\left[\nabla J\left(\theta_{k}\right)\right]_{i}\rvert \times\lvert\mathbb{E}\left\{\frac{\delta \sqrt{1-\beta_{2}} g_{k, i}^{2}}{\sqrt{\zeta_k\left(\delta^{2} \beta_{2} y_{k, i}+\zeta_k\right)}} \mid \theta_{k}\right\}\rvert\right) \\
&+\frac{L \alpha^{2}}{2} \sum_{i=1}^{n} \mathbb{E}\left\{\frac{g_{k, i}^{2}}{\delta^{2} y_{k+1, i}+\zeta_k} \mid \theta_{k}\right\} \\
=&J\left(\theta_{k}\right)-\alpha \sum_{i=1}^{n} \frac{\left[\nabla J\left(\theta_{k}\right)\right]_{i}^{2}}{\sqrt{\delta^{2} \beta_{2} y_{k, i}+\zeta_k}} \\
&+\alpha \sum_{i=1}^{n}\left(\lvert\left[\nabla J\left(\theta_{k}\right)\right]_{i}\rvert \times \mathbb{E}\left\{\frac{\delta \sqrt{1-\beta_{2}} g_{k, i}^{2}}{\sqrt{\zeta_k\left(\delta^{2} \beta_{2} y_{k, i}+\zeta_k\right)}} \mid \theta_{k}\right\}\right) \\
&+\frac{L \alpha^{2}}{2} \sum_{i=1}^{n} \mathbb{E}\left\{\frac{g_{k, i}^{2}}{\delta^{2} y_{k+1, i}+\zeta_k} \mid \theta_{k}\right\}.
\end{aligned}
\end{displaymath}

Using Corollary \ref{corollary_4}, i.e., $\lvert[\nabla J(\theta)]_{i}\rvert \leq M$ for any $i\in\left\{1,2,\cdots,n\right\}$, we have
\begin{displaymath}
\begin{aligned}
&\mathbb{E}\left\{J\left(\theta_{k+1}\right) \mid \theta_{k}\right\} \\
\leq &J\left(\theta_{k}\right)-\alpha \sum_{i=1}^{n} \frac{\left[\nabla J\left(\theta_{k}\right)\right]_{i}^{2}}{\sqrt{\delta^{2} \beta_{2} y_{k, i}+\zeta_k}} \\
&+\frac{\alpha M \delta \sqrt{1-\beta_{2}}}{\sqrt{\zeta_k}} \sum_{i=1}^{n} \mathbb{E}\left\{\frac{g_{k, i}^{2}}{\sqrt{\delta^{2} \beta_{2} y_{k, i}+\zeta_k}} \mid \theta_{k}\right\}\\
&+\frac{L \alpha^{2}}{2} \sum_{i=1}^{n} \mathbb{E}\left\{\frac{g_{k, i}^{2}}{\delta^{2} y_{k+1, i}+\zeta_k} \mid \theta_{k}\right\}\\
\end{aligned}
\end{displaymath}
\begin{displaymath}
\begin{aligned}
\leq &J\left(\theta_{k}\right)-\alpha \sum_{i=1}^{n} \frac{\left[\nabla J\left(\theta_{k}\right)\right]_{i}^{2}}{\sqrt{\delta^{2} \beta_{2} y_{k, i}+\zeta_k}} \\
&+\frac{\alpha M \delta \sqrt{1-\beta_{2}}}{\sqrt{\zeta_k}} \sum_{i=1}^{n} \mathbb{E}\left\{\frac{g_{k, i}^{2}}{\sqrt{\delta^{2} \beta_{2} y_{k, i}+\zeta_k}} \mid \theta_{k}\right\} \\
&+\frac{L \alpha^{2}}{2 \sqrt{\zeta_k}} \sum_{i=1}^{n} \mathbb{E}\left\{\frac{g_{k, i}^{2}}{\sqrt{\delta^{2} y_{k+1, i}+\zeta_k}} \mid \theta_{k}\right\}\\
=&J\left(\theta_{k}\right)-\alpha \sum_{i=1}^{n} \frac{\left[\nabla J\left(\theta_{k}\right)\right]_{i}^{2}}{\sqrt{\delta^{2} \beta_{2} y_{k, i}+\zeta_k}} \\
&+\frac{\alpha M \delta \sqrt{1-\beta_{2}}}{\sqrt{\zeta_k}} \sum_{i=1}^{n} \mathbb{E}\left\{\frac{g_{k, i}^{2}}{\sqrt{\delta^{2} \beta_{2} y_{k, i}+\zeta_k}} \mid \theta_{k}\right\}\\
&+\frac{L \alpha^{2}}{2 \sqrt{\zeta_k}} \sum_{i=1}^{n} \mathbb{E}\left\{\frac{g_{k, i}^{2}}{\sqrt{\delta^{2}\left(\beta_{2} y_{k, i}+\left(1-\beta_{2}\right) g_{k, i}^{2}\right)+\zeta_k}} \mid \theta_{k}\right\}\\
\leq &J\left(\theta_{k}\right)-\alpha \sum_{i=1}^{n} \frac{\left[\nabla J\left(\theta_{k}\right)\right]_{i}^{2}}{\sqrt{\delta^{2} \beta_{2} y_{k, i}+\zeta_k}} \\
&+\frac{\alpha M \delta \sqrt{1-\beta_{2}}}{\sqrt{\zeta_k}} \sum_{i=1}^{n} \mathbb{E}\left\{\frac{g_{k, i}^{2}}{\sqrt{\delta^{2} \beta_{2} y_{k, i}+\zeta_k}} \mid \theta_{k}\right\}\\
&+\frac{L \alpha^{2}}{2 \sqrt{\zeta_k}} \sum_{i=1}^{n} \mathbb{E}\left\{\frac{g_{k, i}^{2}}{\sqrt{\delta^{2} \beta_{2} y_{k, i}+\zeta_k}} \mid \theta_{k}\right\} \\
=&J\left(\theta_{k}\right)-\alpha \sum_{i=1}^{n} \frac{\left[\nabla J\left(\theta_{k}\right)\right]_{i}^{2}}{\sqrt{\delta^{2} \beta_{2} y_{k, i}+\zeta_k}} \\
&+\left(\frac{\alpha M \delta \sqrt{1-\beta_{2}}}{\sqrt{\zeta_k}}+\frac{L \alpha^{2}}{2 \sqrt{\zeta_k}}\right) \sum_{i=1}^{n} \frac{\mathbb{E}\left\{g_{k, i}^{2} \mid \theta_{k}\right\}}{\sqrt{\delta^{2} \beta_{2} y_{k, i}+\zeta_k}}.
\end{aligned}
\end{displaymath}

The last equation is possible because $y_k$ is independent of $\mathcal{B}_k$. Using Lemma \ref{lemma_1}, we have 
\begin{displaymath}
\begin{aligned}
&\mathbb{E}\left\{J\left(\theta_{k+1}\right) \mid \theta_{k}\right\} \\
\leq &J\left(\theta_{k}\right)-\alpha \sum_{i=1}^{n} \frac{\left[\nabla J\left(\theta_{k}\right)\right]_{i}^{2}}{\sqrt{\delta^{2} \beta_{2} y_{k, i}+\zeta_k}} +\left(\frac{\alpha M \delta \sqrt{1-\beta_{2}}}{\sqrt{\zeta_k}}+\frac{L \alpha^{2}}{2 \sqrt{\zeta_k}}\right) \\
&\times\sum_{i=1}^{n} \frac{1}{\sqrt{\delta^{2} \beta_{2} y_{k, i}+\zeta_k}}\left(\frac{\sigma_{i}^{2}}{B_{k}}+\left[\nabla J\left(\theta_{k}\right)\right]_{i}^{2}\right) \\
=&J\left(\theta_{k}\right)\\
&+\left(-\alpha+\frac{\alpha M \delta \sqrt{1-\beta_{2}}}{\sqrt{\zeta_k}}+\frac{L \alpha^{2}}{2 \sqrt{\zeta_k}}\right) \sum_{i=1}^{n} \frac{\left[\nabla J\left(\theta_{k}\right)\right]_{i}^{2}}{\sqrt{\delta^{2} \beta_{2} y_{k, i}+\zeta_k}} \\
&+\left(\frac{\alpha M \delta \sqrt{1-\beta_{2}}}{\sqrt{\zeta_k}}+\frac{L \alpha^{2}}{2 \sqrt{\zeta_k}}\right) \sum_{i=1}^{n} \frac{\sigma_{i}^{2}}{B_{k} \sqrt{\delta^{2} \beta_{2} y_{k, i}+\zeta_k}}.
\end{aligned}
\end{displaymath}

If we choose $\alpha \leq \frac{\sqrt{\zeta_{0}}}{2 L}$ and $\beta_{2} \geq 1-\frac{\zeta_{0}}{16 M^{2} \delta^{2}}$, thus having the following conditions:
\begin{displaymath}
\begin{aligned}
\frac{L \alpha}{2 \sqrt{\zeta_k}} & \leq \frac{L \alpha}{2 \sqrt{\zeta_{0}}} \leq \frac{1}{4}, \\
\frac{M \delta \sqrt{1-\beta_{2}}}{\sqrt{\zeta_k}} & \leq \frac{M \delta \sqrt{1-\beta_{2}}}{\sqrt{\zeta_{0}}} \leq \frac{1}{4}.
\end{aligned}
\end{displaymath}

Using these inequations and Corollary \ref{corollary_5}, we obtain
\begin{displaymath}
\begin{aligned}
&\mathbb{E}\left\{J\left(\theta_{k+1}\right) \mid \theta_{k}\right\} \\
\leq & J\left(\theta_{k}\right)-\frac{\alpha}{2} \sum_{i=1}^{n} \frac{\left[\nabla J\left(\theta_{k}\right)\right]_{i}^{2}}{\sqrt{\delta^{2} \beta_{2} y_{k, i}+\zeta_k}} \\
&+\left(\frac{\alpha M \delta \sqrt{1-\beta_{2}}}{\sqrt{\zeta_k}}+\frac{L \alpha^{2}}{2 \sqrt{\zeta_k}}\right) \sum_{i=1}^{n} \frac{\sigma_{i}^{2}}{B_{k} \sqrt{\delta^{2} \beta_{2} y_{k, i}+\zeta_k}} \\
\leq & J\left(\theta_{k}\right)-\frac{\alpha}{2} \frac{\left\|\nabla J\left(\theta_{k}\right)\right\|^{2}}{\sqrt{\delta^{2} \beta_{2} M^{2}+\zeta_k}}\\
&+\left(\frac{\alpha M \delta \sqrt{1-\beta_{2}}}{\sqrt{\zeta_k}}+\frac{L \alpha^{2}}{2 \sqrt{\zeta_k}}\right) \frac{\sigma^{2}}{B_{k} \sqrt{\zeta_k}} \\
\leq & J\left(\theta_{k}\right)-\frac{\alpha}{2} \frac{\left\|\nabla J\left(\theta_{k}\right)\right\|^{2}}{\sqrt{\delta^{2} \beta_{2} M^{2}+\zeta_k}}+\left(\frac{\alpha}{4}+\frac{\alpha}{4}\right) \frac{\sigma^{2}}{B_{k} \sqrt{\zeta_k}} \\
\leq & J\left(\theta_{k}\right)-\frac{\alpha}{2} \frac{\left\|\nabla J\left(\theta_{k}\right)\right\|^{2}}{\sqrt{\delta^{2} \beta_{2} M^{2}+1}}+\frac{\alpha \sigma^{2}}{2 B_{k} \sqrt{\zeta_k}} .
\end{aligned}
\end{displaymath}

The second inequation follows from the fact that $0\leq y_{k,i}\leq M^2$ and $\sqrt{\delta^{2} \beta_{2} y_{k, i}+\zeta_k} \geq \sqrt{\zeta_k}$. The fourth inequation follows from the fact that $0<\zeta_k\leq1$. We further take expectations of all variables on both sides and add up all the inequations from $k=0$ to $k=N-1$ and obtain
\begin{displaymath}
\begin{aligned}
&\frac{\alpha}{2 \sqrt{\delta^{2} \beta_{2} M^{2}+1}} \sum_{k=0}^{N-1} \mathbb{E}\left\{\left\|\nabla J\left(\theta_{k}\right)\right\|^{2}\right\} \\
\leq &\sum_{k=0}^{N-1}\left(\mathbb{E}\left\{J\left(\theta_{k}\right)\right\}-\mathbb{E}\left\{J\left(\theta_{k+1}\right)\right\}\right) +\frac{\alpha}{2} \sum_{k=0}^{N-1} \frac{\sigma^{2}}{B_{k}\sqrt{\zeta_k}} \\
=&J\left(\theta_{0}\right)-\mathbb{E}\left\{J\left(\theta_{N}\right)\right\}+\frac{\alpha}{2} \sum_{k=0}^{N-1} \frac{\sigma^{2}}{B_{k} \sqrt{\zeta_k}} .
\end{aligned}
\end{displaymath}

Multiply with $\nicefrac{2 \sqrt{\delta^{2} \beta_{2} M^{2}+1}}{\alpha N}$ on both sides and using the fact that $J\left(\theta^{*}\right) \leq \mathbb{E}\left\{J\left(\theta_{N}\right)\right\}$, we obtain the following bound:
\begin{displaymath}
\begin{aligned}
&\frac{1}{N} \sum_{k=0}^{N-1} \mathbb{E}\left\{\left\|\nabla J\left(\theta_{k}\right)\right\|^{2}\right\} \\
\leq& 2 \sqrt{\delta^{2} \beta_{2} M^{2}+1} \left(\frac{J\left(\theta_{0}\right)-J\left(\theta^{*}\right)}{\alpha N}+\frac{\sigma^{2}}{2 N} \sum_{k=0}^{N-1} \frac{1}{B_{k}\sqrt{\zeta_k}}\right).
\end{aligned}
\end{displaymath}

\bibliographystyle{IEEEtran}
\bibliography{cite}


\begin{IEEEbiography}[{\includegraphics[width=1in,height=1.25in,clip,keepaspectratio]{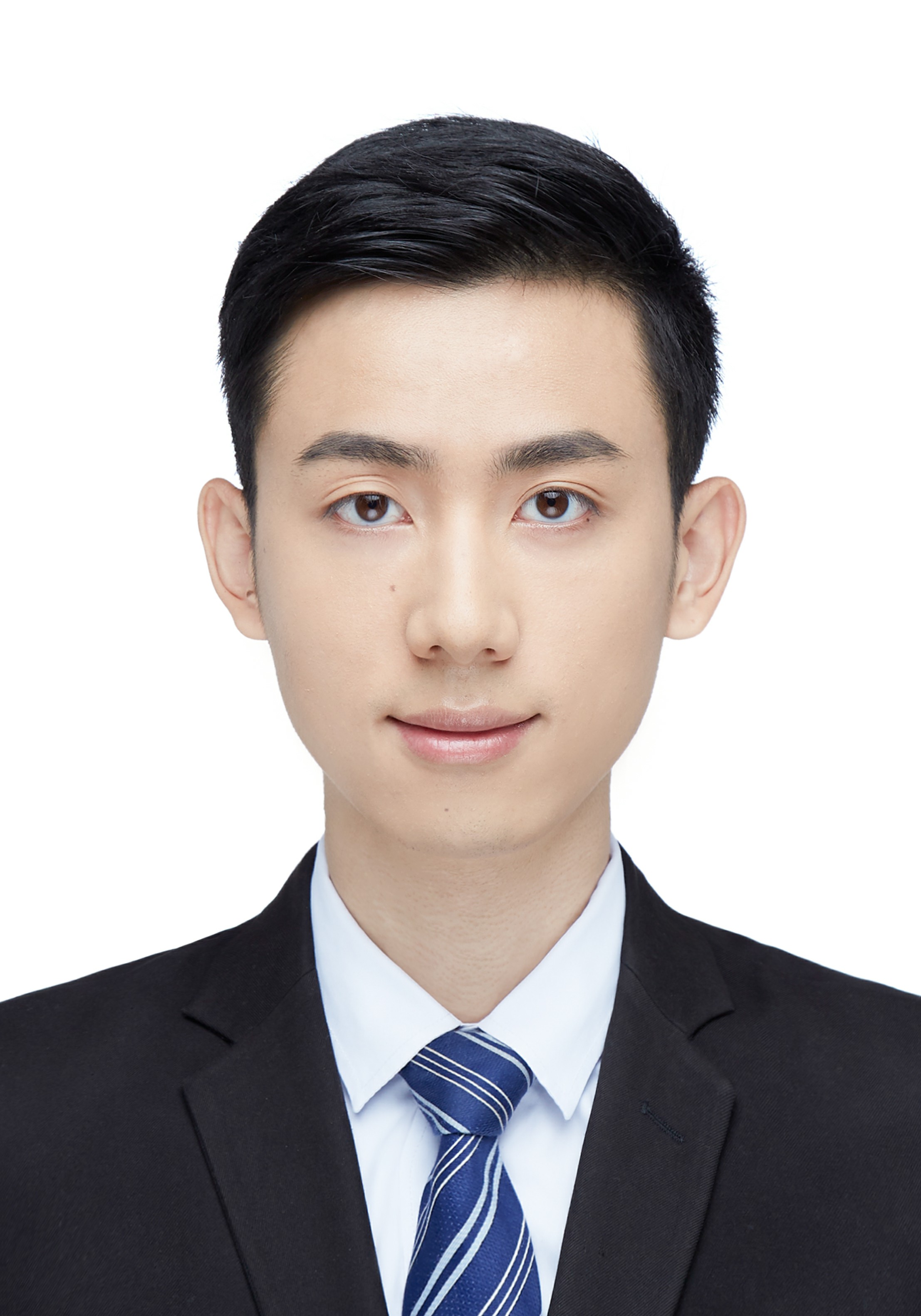}}]{Yao Lyu}
received his B.S. degree in Automotive Engineering from Tsinghua University, Beijing, China, in 2019. He is currently pursuing his Ph.D. degree in Automotive Engineering from School of Vehicle and Mobility, Tsinghua University, Beijing, China. His active research interests include decision-making and control of autonomous vehicles, reinforcement learning, accelerating optimization and optimal control theory.
\end{IEEEbiography}

\begin{IEEEbiography}[{\includegraphics[width=1in,height=1.25in,clip,keepaspectratio]{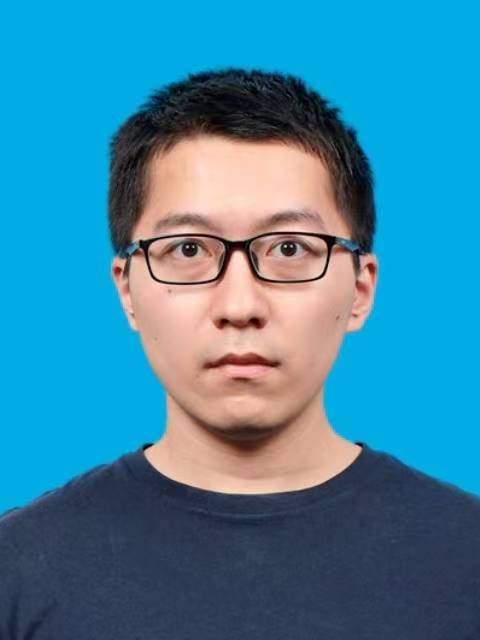}}]{Xiangteng Zhang}
received the B.S. degree in engineering mechanics from the School of Aerospace Engineering, Tsinghua University, Beijing, China, in 2022. He is currently working toward the Ph.D. degree in mechanical engineering with the School of Vehicle and Mobility, Tsinghua University, Beijing. His current research interests include optimal control theory, reinforcement learning, and decision and control of autonomous vehicles.
\end{IEEEbiography}

\begin{IEEEbiography}[{\includegraphics[width=1in,height=1.25in,clip,keepaspectratio]{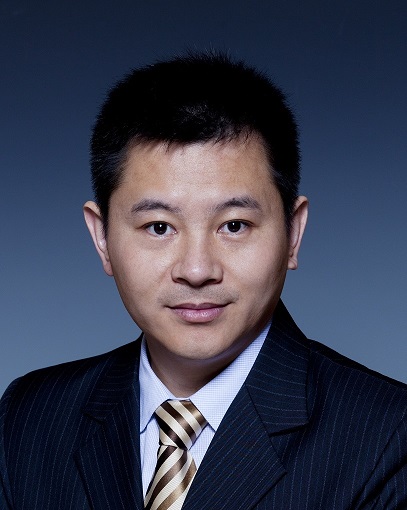}}]{Shengbo Eben Li}
(Senior Member, IEEE) received his M.S. and Ph.D. degrees from Tsinghua University in 2006 and 2009. He is currently a Tenured Professor with Tsinghua University. He has published more than 130 peer-reviewed papers in top-tier international journals and conferences. He is a recipient of the National Award for Technological Invention of China (2013), National Award for Progress in Sci \& Tech of China (2018), Distinguished Young Scholar of Beijing NSF (2018), etc. His research interests include intelligent vehicles and driver assistance, deep reinforcement learning, optimal control and estimation. 
\end{IEEEbiography}

\begin{IEEEbiography}[{\includegraphics[width=1in,height=1.25in,clip,keepaspectratio]{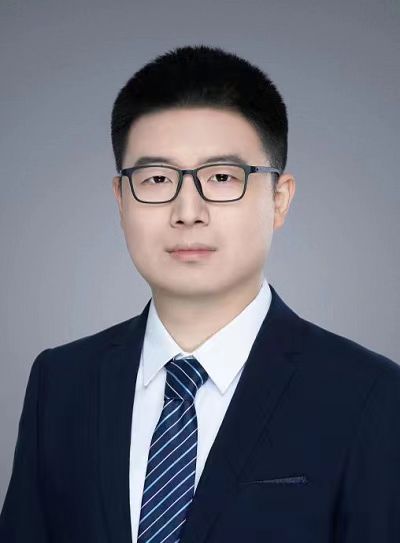}}]{Jingliang Duan}
(Member, IEEE) received his Ph.D. degree in mechanical engineering from Tsinghua University, Beijing, China, in 2021. 
Following his Ph.D., he served as a research fellow in the Department of Electrical and Computer Engineering at the National University of Singapore from 2021 to 2022. He is currently a tenured associate professor in the School of Mechanical Engineering, University of Science and Technology Beijing, China. His research interests include reinforcement learning, optimal control, and self-driving decision-making.
\end{IEEEbiography}



\begin{IEEEbiography}[{\includegraphics[width=1in,height=1.25in,clip,keepaspectratio]{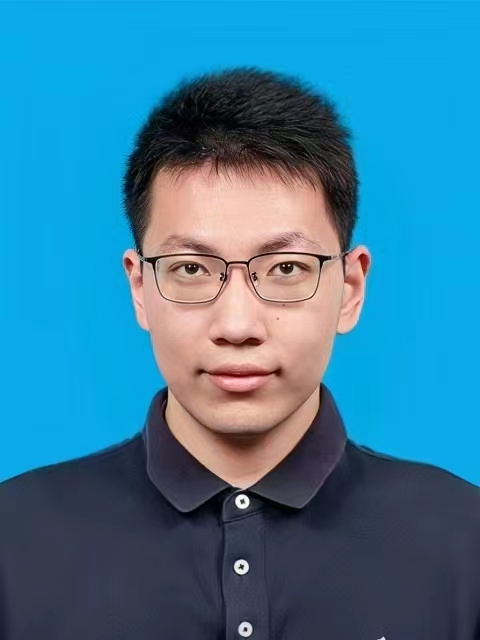}}]{Letian Tao}
received the B.S. degree in Vehicle Engineering from the School of Vehicle and Mobility, Tsinghua University, Beijing, in 2022. He is currently pursuing the Ph.D. degree in Mechanical Engineering at the School of Vehicle and Mobility, Tsinghua University, Beijing. His research interests focus on reinforcement learning and optimal control, particularly their applications in autonomous driving.
\end{IEEEbiography}

\begin{IEEEbiography}[{\includegraphics[width=1in,height=1.25in,clip,keepaspectratio]{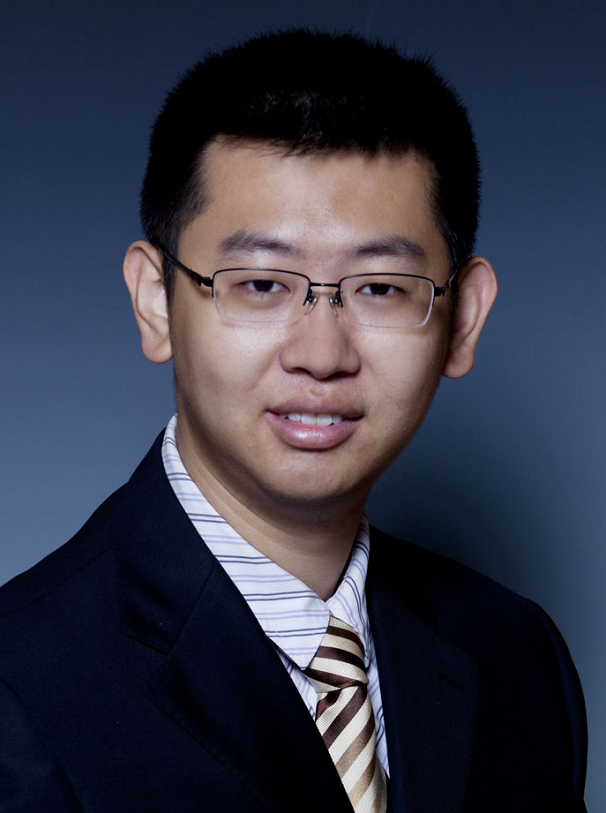}}]{Qing Xu}
received the B.S., M.S., and Ph.D. degrees in automotive engineering from Beihang University, Beijing, China, in 2006, 2008, and 2014, respectively. During his Ph.D. research, he worked as a Visiting Scholar with the Department of Mechanical Science and Engineering, University of Illinois at Urbana-Champaign, Champaign, IL, USA. From 2014 to 2016, he had his post-doctoral research at Tsinghua University, Beijing, where he is currently working as an Associate Research Professor with the School of Vehicle and Mobility. His main research interests include the decision and control of intelligent vehicles.
\end{IEEEbiography}

\begin{IEEEbiography}[{\includegraphics[width=1in,height=1.25in,clip,keepaspectratio]{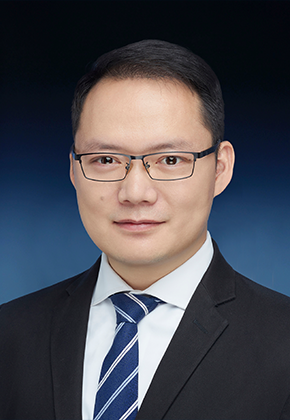}}]{Lei He} received the B.S. degree in aeronautical engineering from Beijing University of Aeronautics and Astronautics, Beijing, China, in 2013, and the Ph.D. degree from the Institute of Automation, Chinese Academy of Sciences, Beijing, in 2018. 
He is currently an Assistant Research Fellow with the School of Vehicle and Mobility, Tsinghua University. His research interests include computer vision, machine learning, and autonomous driving. Dr. He has authored over 40 granted patents and multiple academic papers in international journals and conferences such as IEEE TIP and ICRA. 
\end{IEEEbiography}

\begin{IEEEbiography}[{\includegraphics[width=1in,height=1.25in,clip,keepaspectratio]{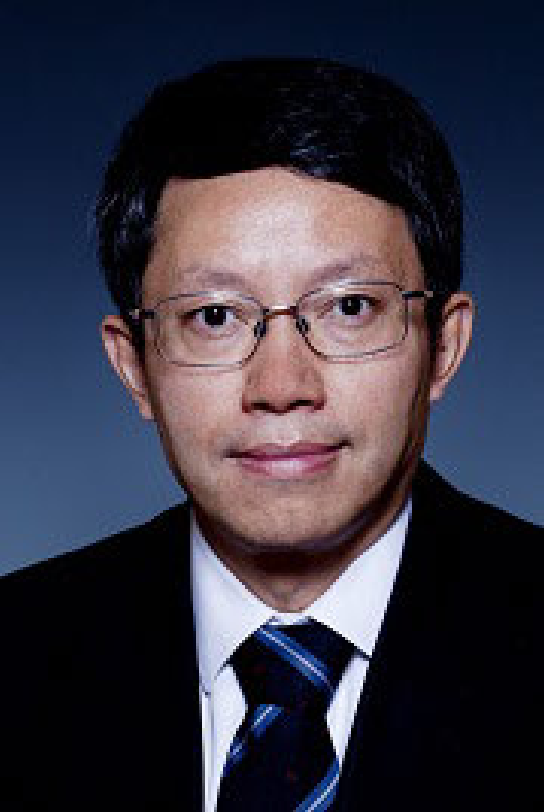}}]{Keqiang Li}
received the B.Tech. degree from Tsinghua University, Beijing, China, in 1985, and the M.S. and Ph.D. degrees in mechanical engineering from the Chongqing University of China, Chongqing, China, in 1988 and 1995, respectively. He is currently a Professor with the School of Vehicle and Mobility, Tsinghua University. He has authored more than 200 articles and is a co-inventor of over 80 patents in China and Japan. Dr. Li served as a member for the Chinese Academy of Engineering. His main research areas include automotive control systems, driver assistance systems, and networked dynamics and control. 
\end{IEEEbiography}


\clearpage
\setcounter{page}{1}
\begin{onecolumn}
\renewcommand{\appendixname}{}
\appendix
\addcontentsline{toc}{section}{Supplementary material}

\begin{center}
{\Huge Supplementary Material for \textit{Conformal Symplectic Optimization for Stable Reinforcement Learning\\}}
\vskip 10pt
{Yao Lyu, Xiangteng Zhang, Shengbo Eben Li*, Jingliang Duan, Letian Tao, Qing Xu, Lei He, Keqiang Li}
\end{center}

\section*{S.I Basis of conformal symplectic optimization}
\addcontentsline{toc}{section}{Basis of conformal symplectic optimization}
This section introduces some basis of conformal symplectic optimization, including symplectic integrators and splitting methods.

\subsection{Symplectic integrators}
A one-step numerical method is called symplectic if the one-step map:
\begin{displaymath}
z_{1}=\phi_{h}\left(z_{0}\right)
\end{displaymath}
is symplectic whenever the method is applied to a smooth, conservative Hamiltonian system. The following theorems show two typical symplectic integrators with respect to the conservative Hamiltonian system $\dot{z}=S \nabla H(z)=\mathcal{C}(z)$ \cite{hairer2006geometric}.
\begin{theorem}
The so-called symplectic Euler methods:
\begin{equation}
\label{eq_sym_euler_1}
p_{k+1}=p_{k}-h \nabla_{q} H\left(q_{k}, p_{k+1}\right), \quad
q_{k+1}=q_{k}+h \nabla_{p} H\left(q_{k}, p_{k+1}\right),
\end{equation}
or
\begin{equation}
\label{eq_sym_euler_2}
p_{k+1}=p_{k}-h \nabla_{q} H\left(q_{k+1}, p_{k}\right), \quad
q_{k+1}=q_{k}+h \nabla_{p} H\left(q_{k+1}, p_{k}\right),
\end{equation}
are symplectic methods of order 1. 
\end{theorem}

\begin{theorem}
The Verlet schemes, i.e., leapfrog methods:
\begin{equation}
\label{eq_leapfrog_1}
\begin{aligned}
p_{k+\frac{1}{2}}&=p_{k}-\frac{h}{2} \nabla_{q} H\left(q_{k}, p_{k+\frac{1}{2}}\right), \\
q_{k+1}&=q_{k}+\frac{h}{2}\left(\nabla_{p} H\left(q_{k}, p_{k+\frac{1}{2}}\right)+\nabla_{p} H\left(q_{k+1}, p_{k+\frac{1}{2}}\right)\right), \\
p_{k+1}&=p_{k+\frac{1}{2}}-\frac{h}{2} \nabla_{q} H\left(q_{k+1}, p_{k+\frac{1}{2}}\right),
\end{aligned}
\end{equation}
or
\begin{equation}
\label{eq_leapfrog_2}
\begin{aligned}
q_{k+\frac{1}{2}}&=q_{k}+\frac{h}{2} \nabla_{p} H\left(q_{k+\frac{1}{2}}, p_{k}\right), \\
p_{k+1}&=p_{k}-\frac{h}{2}\left(\nabla_{q} H\left(q_{k+\frac{1}{2}}, p_{k}\right)+\nabla_{q} H\left(q_{k+\frac{1}{2}}, p_{k+1}\right)\right), \\
q_{k+1}&=q_{k+\frac{1}{2}}+\frac{h}{2} \nabla_{p} H\left(q_{k+\frac{1}{2}}, p_{k+1}\right),
\end{aligned}
\end{equation}
are symplectic methods of order 2.
\end{theorem}

\subsection{Splitting methods}
\begin{definition}[Splitting methods]
Consider an arbitrary system $\dot{z}=\zeta(z)$ in $\mathbb{R}^{2n}$, and suppose that the vector field is “split” as
\begin{displaymath}
\dot{z}=\zeta^{[1]}(z)+\zeta^{[2]}(z).
\end{displaymath}

If the exact flow $\varphi_{t}^{[1]}$ and $\varphi_{t}^{[2]}$ of the system $\dot{z}=\zeta^{[1]}(z)$ and $\dot{z}=\zeta^{[2]}(z)$ can be calculated explicitly, we can compose their numerical maps $\phi_{h}^{[1]}$ and $\phi_{h}^{[2]}$ to get the numerical approximation $\phi_{h}=\phi_{h}^{[1]} \circ \phi_{h}^{[2]}$ of the system $\dot{z}=\zeta(z)$.
\end{definition}
Moreover, it is proved that splitting methods in the composition $\phi_{h}^{[1]} \circ \phi_{h}^{[2]}$ produces a first-order integrator, while outputs a second-order integrator in the composition $\phi_{\nicefrac{h}{2}}^{[2]} \circ \phi_{h}^{[1]} \circ \phi_{\nicefrac{h}{2}}^{[2]}$ \cite{hairer2006geometric}.

\subsection{Derivative of the DLPF algorithm}
\label{appendix_HB & LPF}
This section provides a detailed analysis of the derivative of the DLPF algorithm through the discretization of conformal Hamiltonian systems comprising numerous independent microscopic one-dimensional particles. If we choose the leapfrog method \eqref{eq_leapfrog_2} to integrate the conservative flow $\varphi^{\mathcal{C}}_t$ and consider the composition $\phi_{h}=\phi_{h / 2}^{\mathcal{D}} \circ \phi_{h}^{\mathcal{C}} \circ \phi_{h / 2}^{\mathcal{D}}$, we obtain
\begin{equation}
\label{eq_step3_2}
\begin{aligned}
q_{k+\frac{1}{2}}&=q_{k}+\frac{h}{2} \nabla_{p} H\left(q_{k+\frac{1}{2}}, e^{-\frac{1}{2} r h} p_{k}\right), \\
p_{k+\frac{1}{2}}&=e^{-\frac{1}{2} r h} p_{k}-\frac{h}{2}\left(\nabla_{q} H\left(q_{k+\frac{1}{2}}, e^{-\frac{1}{2} r h} p_{k}\right)+\nabla_{q} H\left(q_{k+\frac{1}{2}}, p_{k+\frac{1}{2}}\right)\right), \\
q_{k+1}&=q_{k+\frac{1}{2}}+\frac{h}{2} \nabla_{p} H\left(q_{k+\frac{1}{2}}, p_{k+\frac{1}{2}}\right), \\
p_{k+1}&=e^{-\frac{1}{2} r h} p_{k+\frac{1}{2}}.
\end{aligned}
\end{equation}

Then, let us consider the classical Hamiltonian with each one-dimensional particle possessing the same mass $m$, i.e., $H(\theta,p)=\sum_{i=1}^n \frac{p_i^2}{2m}+J(\theta)$, and replace it into \eqref{eq_step3_2}. Introducing the change of variables \eqref{eq_variable_change_1}, we have
\begin{equation}
\label{eq_dis_lpf_before_rearranging}
\begin{aligned}
\theta_{k+\frac{1}{2}}&=\theta_{k}-\frac{1}{2} \alpha \sqrt{\beta_1} v_{k}, \\
v_{k+\frac{1}{2}}&=\sqrt{\beta_1} v_{k}+(1-\beta_1) \nabla J\left(\theta_{k+\frac{1}{2}}\right), \\
\theta_{k+1}&=\theta_{k+\frac{1}{2}}-\frac{1}{2} \alpha v_{k+\frac{1}{2}}, \\
v_{k+1}&=\sqrt{\beta_1} v_{k+\frac{1}{2}}.
\end{aligned}
\end{equation}

Note that the algorithm remains the same if we replace successive updating rules. Thus, we rewrite them into two-step updating rules and obtain
\begin{equation}
\label{eq_dis_lpf}
\begin{aligned}
v_{k+\frac{1}{2}} &=\sqrt{\beta_1} \cdot \sqrt{\beta_1} v_{k-\frac{1}{2}}+(1-\beta_1) \nabla J\left(\theta_{k+\frac{1}{2}}\right) \\
&=\beta_1 v_{k-\frac{1}{2}}+(1-\beta_1) \nabla J\left(\theta_{k+\frac{1}{2}}\right), \\
\theta_{k+\frac{3}{2}} &=\theta_{k+1}-\frac{1}{2} \alpha \sqrt{\beta_1} v_{k+1} \\
&=\theta_{k+\frac{1}{2}}-\frac{1}{2} \alpha v_{k+\frac{1}{2}}-\frac{1}{2} \alpha \sqrt{\beta_1} \cdot \sqrt{\beta_1} v_{k+\frac{1}{2}} \\
&=\theta_{k+\frac{1}{2}}-\frac{1}{2} \alpha(\beta_1+1) v_{k+\frac{1}{2}}.
\end{aligned}
\end{equation}

The DLPF algorithm is a second-order conformal symplectic integrator \cite{francca2020conformal}, whose pseudocode is shown in Algorithm \ref{algorithm_dis_lpf}. It is worth noting that $\theta_k$ and $v_k$ in Algorithm \ref{algorithm_dis_lpf} actually means $\theta_{k+\frac{1}{2}}$ and $v_{k-\frac{1}{2}}$ in \eqref{eq_dis_lpf}, respectively. But the only distinction is that their subscripts are different, which is written on purpose for the convenience and unification of the pseudocode.

\begin{algorithm}
\caption{Dissipative leapfrog (DLPF) algorithm \cite{francca2021dissipative}}
\label{algorithm_dis_lpf}
\renewcommand{\algorithmicrequire}{\textbf{Input:}}
\begin{algorithmic}[1]
\REQUIRE parameters of neural network $\theta_0$ and their conjugate momenta $v_0$, learning rate $\alpha>0$, first-order momentum coefficient $0<\beta_1<1$
\FOR{$k=0$ {\bf to} $N-1$}
    \STATE $v_{k+1}=\beta_1 v_{k}+(1-\beta_1) \nabla J\left(\theta_{k}\right)$ 
    \STATE $g_{k}=\frac{1}{2}(\beta_1+1) v_{k+1}$ 
    \STATE $\theta_{k+1}=\theta_{k}-\alpha g_{k}$
\ENDFOR
\end{algorithmic}
\end{algorithm}

\clearpage
\section*{S.II More details on RAD}
\addcontentsline{toc}{section}{More details on RAD}
\setcounter{subsection}{0}
\subsection{Relativistic Hamiltonian}
\label{appendix_relativistic Hamiltonian}
Here, we derive the relativistic Hamiltonian in the form of \eqref{eq_relativistic_H}. In special relativity, the well-known mass-energy equation is
\begin{equation}
\nonumber
E=m \gamma_r c^2,
\end{equation}
where $E$ is the total energy of a single particle, $m$ is the rest mass of the single particle, $\gamma_r=\frac{1}{\sqrt{1-s^2/c^2}}$ is the relativistic coefficient, $s$ is the speed of the single particle, and $c$ is the speed of light. Therefore, the rest energy, in which case the $s=0$, is
\begin{equation}
\nonumber
E_0=m c^2,
\end{equation}
and the momentum of a single particle is
\begin{equation}
\nonumber
p_i=m \gamma_r s.
\end{equation}

Here, it is easy to obtain the following equation:
\begin{equation}
\nonumber
c^2p_i^2+E_0^2=\frac{m^2c^4}{1-s^2/c^2}=E^2,
\end{equation}

Hence, the kinetic energy of the single particle is
\begin{equation}
\nonumber
T(p_i)=E-E_0=\sqrt{c^2p_i^2+E_0^2}-E_0=c\sqrt{p_i^2+m^2c^2}-E_0.
\end{equation}

Since the rest energy $E_0$ is constant, having no contribution to the system's canonical equations, we can ignore it and have the Hamiltonian
\begin{equation}
\nonumber
H(q,p)=T(p)+U(q)=\sum_i T(p_i)+U(q)=\sum_i c\sqrt{p_i^2+m^2c^2}+U(q).
\end{equation}

Finally, we obtain \eqref{eq_relativistic_H} by replacing the potential energy $U(q)$ with the objective $J(\theta)$.

\subsection{Derivative of the second-order RAD}
\label{appendix_derivative of the second-order RAD}
Consider the relativistic Hamiltonian \eqref{eq_relativistic_H} and replace it into \eqref{eq_step3_2}, we receive the following integrator:
\begin{equation}
\nonumber
\label{eq_rad_2_original_1}
\begin{aligned}
\theta_{k+\frac{1}{2}}&=\theta_{k}+\frac{h c}{2} \frac{e^{-\frac{1}{2} r h}}{\sqrt{e^{-r h} p_{k}^{2}+m^{2} c^{2}}} p_{k}, \\
p_{k+\frac{1}{2}}&=e^{-\frac{1}{2} r h} p_{k}-h \nabla J\left(\theta_{k+\frac{1}{2}}\right), \\
\theta_{k+1}&=\theta_{k+\frac{1}{2}}+\frac{h c}{2} \frac{1}{\sqrt{p_{k+\frac{1}{2}}^{2}+m^{2} c^{2}}} p_{k+\frac{1}{2}}, \\
p_{k+1}&=e^{-\frac{1}{2} r h} p_{k+\frac{1}{2}}.
\end{aligned}
\end{equation}

Introducing the changes of variables \eqref{eq_variable_change_1}, thus we obtain
\begin{equation}
\nonumber
\label{eq_rad_2_original_before_rearranging}
\begin{aligned}
\theta_{k+\frac{1}{2}}&=\theta_{k}-\frac{\alpha / 2}{\sqrt{\delta^{2} v_{k}^{2}+\frac{1}{\beta_1}}} v_{k}, \\
v_{k+\frac{1}{2}}&=\sqrt{\beta_1} v_{k}+(1-\beta_1) \nabla J\left(\theta_{k+\frac{1}{2}}\right), \\
\theta_{k+1}&=\theta_{k+\frac{1}{2}}-\frac{\alpha / 2}{\sqrt{\delta^{2} v_{k+\frac{1}{2}}^{2}+1}} v_{k+\frac{1}{2}}, \\
v_{k+1}&=\sqrt{\beta_1} v_{k+\frac{1}{2}}.
\end{aligned}
\end{equation}

Since replacing successive updating rules does not change the algorithm, rewriting them into two-step updating rules as
\begin{equation}
\nonumber
\label{eq_rad_2_original}
\begin{aligned}
v_{k+\frac{1}{2}}&=\sqrt{\beta_1} \cdot \sqrt{\beta_1} v_{k-\frac{1}{2}}+(1-\beta_1) \nabla J\left(\theta_{k+\frac{1}{2}}\right)\\
&=\beta_1 v_{k-\frac{1}{2}}+(1-\beta_1) \nabla J\left(\theta_{k+\frac{1}{2}}\right) \text {, }\\
\theta_{k+\frac{3}{2}}&=\theta_{k+1}-\frac{\alpha / 2}{\sqrt{\delta^{2} v_{k+1}^{2}+\frac{1}{\beta_1}}} v_{k+1}\\
&=\theta_{k+\frac{1}{2}}-\frac{\alpha / 2}{\sqrt{\delta^{2} v_{k+\frac{1}{2}}^{2}+1}} v_{k+\frac{1}{2}}-\frac{\alpha / 2}{\sqrt{\delta^{2} \beta_1 v_{k+\frac{1}{2}}^{2}+\frac{1}{\beta_1}}} \sqrt{\beta_1} v_{k+\frac{1}{2}}\\
&=\theta_{k+\frac{1}{2}}-\left(\frac{1}{\sqrt{\delta_{k+\frac{1}{2}}^{2}+1}}+\frac{1}{\sqrt{\delta^{2} v_{k+\frac{1}{2}}^{2}+\frac{1}{\beta_1^{2}}}}\right) \frac{\alpha}{2} v_{k+\frac{1}{2}},
\end{aligned}
\end{equation}
and we immediately receive the original form of the second-order RAD related to the relativistic Hamiltonian \eqref{eq_relativistic_H}.

\clearpage
\section*{S.III Basis of other optimization algorithms}
\label{appendix_basis of other integrators}
\addcontentsline{toc}{section}{Basis of other optimization algorithms}
This section introduces some basis of the other integrators involved in this paper, including SGD, NAG and ADAM.

\setcounter{subsection}{0}
\subsection{Stochastic gradient descent algorithm}
SGD is precisely a forward Euler discretization for the differential equation $\dot{\theta}=-\nabla J(\theta)$ \cite{yazan2017comparison}, and its pseudocode is shown in Algorithm \ref{algorithm_SGD}.

\begin{algorithm}
\caption{Stochastic gradient descent (SGD) algorithm \cite{yazan2017comparison}}
\label{algorithm_SGD}
\renewcommand{\algorithmicrequire}{\textbf{Input:}}
\begin{algorithmic}[1]
\REQUIRE parameters of neural network $\theta_0$, learning rate $\alpha>0$
\FOR{$k=0$ {\bf to} $N-1$}
    \STATE $g_{k}=\nabla J\left(\theta_{k}\right)$
    \STATE $\theta_{k+1}=\theta_{k}-\alpha g_{k}$
\ENDFOR
\end{algorithmic}
\end{algorithm}

\subsection{Nesterov accelerated gradient algorithm}
It is worth noting that there is a close similarity between DLPF and NAG \cite{bottou2018optimization}. DLPF would be exactly NAG if we replace $\theta_{k+\frac{1}{2}}$ with $\theta_k$ in the third equation of \eqref{eq_dis_lpf_before_rearranging} as
\begin{equation}
\nonumber
\label{eq_NAG_before_rearranging}
\begin{aligned}
\theta_{k+\frac{1}{2}}&=\theta_{k}-\frac{1}{2} \alpha \sqrt{\beta_1} v_{k}, \\
v_{k+\frac{1}{2}}&=\sqrt{\beta_1} v_{k}+(1-\beta_1) \nabla J\left(\theta_{k+\frac{1}{2}}\right), \\
\theta_{k+1}&=\theta_{k}-\frac{1}{2} \alpha v_{k+\frac{1}{2}}, \\
v_{k+1}&=\sqrt{\beta_1} v_{k+\frac{1}{2}}.
\end{aligned}
\end{equation}

Rewriting them into two-step updating rules like DLPF, we have
\begin{equation}
\nonumber
\begin{aligned}
v_{k+\frac{1}{2}} &=\sqrt{\beta_1} \cdot \sqrt{\beta_1} v_{k-\frac{1}{2}}+(1-\beta_1) \nabla J\left(\theta_{k+\frac{1}{2}}\right) \\
&=\beta_1 v_{k-\frac{1}{2}}+(1-\beta_1) \nabla J\left(\theta_{k+\frac{1}{2}}\right), \\
\theta_{k+\frac{3}{2}} &=\theta_{k+1}-\frac{1}{2} \alpha \sqrt{\beta_1} v_{k+1} \\
&=\theta_{k}-\frac{1}{2} \alpha v_{k+\frac{1}{2}}-\frac{1}{2} \alpha \sqrt{\beta_1}\cdot\sqrt{\beta_1} v_{k+\frac{1}{2}} \\
&=\theta_{k+\frac{1}{2}}+\frac{1}{2} \alpha\sqrt{\beta_1} v_{k} -\frac{1}{2}\alpha v_{k+\frac{1}{2}} -\frac{1}{2} \alpha \beta_1 v_{k+\frac{1}{2}} \\
&=\theta_{k+\frac{1}{2}}-\frac{1}{2} \alpha(1-\beta_1) \nabla J\left(\theta_{k+\frac{1}{2}}\right)-\frac{1}{2} \alpha \beta_1 v_{k+\frac{1}{2}} \\
& =\theta_{k+\frac{1}{2}}-\frac{1}{2} \alpha\left(\beta_1 v_{k+\frac{1}{2}}+(1-\beta_1) \nabla J\left(\theta_{k+\frac{1}{2}}\right)\right).
\end{aligned}
\end{equation}

This is the famous NAG algorithm, whose pseudocode is shown in Algorithm \ref{algorithm_NAG}. Note that we adjust the subscripts for the convenience and unification of the pseudocode as the same in Algorithm \ref{algorithm_dis_lpf}. Intuitively, pulling the updating base backwards, i.e., replacing $\theta_{k+\frac{1}{2}}$ with $\theta_k$ in the third updating rule of \eqref{eq_dis_lpf_before_rearranging}, introduces unreal dissipation into the original dynamical system \cite{francca2020conformal}. Therefore, although NAG is a first-order integrator of the classical Hamiltonian system, it is not conformal symplectic. Indeed, while DLPF exactly preserves the same dissipation of the continuous-time system, NAG introduces some extra contraction or expansion to the symplectic form, thus changing the behavior of the original system slightly.

\begin{algorithm}
\caption{Nesterov accelerated gradient (NAG) algorithm \cite{bottou2018optimization}}
\label{algorithm_NAG}
\renewcommand{\algorithmicrequire}{\textbf{Input:}}
\begin{algorithmic}[1]
\REQUIRE parameters of neural network $\theta_0$, learning rate $\alpha>0$ 
\FOR{$k=0$ {\bf to} $N-1$}
    \STATE $v_{k+1}=\beta_1 v_{k}+(1-\beta_1) \nabla J\left(\theta_{k}\right)$
    \STATE $g_{k}=\frac{1}{2}\left(\beta_1 v_{k+1}+(1-\beta_1) \nabla J\left(\theta_{k}\right)\right)$
    \STATE $\theta_{k+1}=\theta_{k}-\alpha g_{k}$
\ENDFOR
\end{algorithmic}
\end{algorithm}

\subsection{Adaptive moment gradient algorithm}
\label{appendix_adam}
ADAM is hailed as the most promising algorithm for stochastic nonconvex optimization \cite{zaheer2018adaptive}. It has been utilized in many DL problems and has proven experimentally effective. This algorithm estimates the gradients' first-order and secondary raw moments, then individually adjusts the effective learning rates of different parameters. Its pseudocode from the original paper is shown in Algorithm \ref{algorithm_adam_original}, wherein every operation on vector is element-wise \cite{kingma2014adam}.

\begin{algorithm}
\caption{Adaptive moment estimation-original (ADAM) algorithm (original) \cite{zaheer2018adaptive}}
\label{algorithm_adam_original}
\renewcommand{\algorithmicrequire}{\textbf{Input:}}
\begin{algorithmic}[1]
\REQUIRE parameters of neural network $\theta_0$, first-order momenta $v_0$, second-order momenta $y_0$, learning rate $\alpha>0$, first-order momentum coefficient $0<\beta_1<1$, second-order momentum coefficient $0<\beta_2<1$, rational factor $\hat{\epsilon}>0$
\FOR{$k=0$ {\bf to} $N-1$}
    \STATE $v_{k+1}=\beta_{1} v_{k}+\left(1-\beta_{1}\right) \nabla J\left(\theta_{k}\right) \text { (Update biased first-order moment estimate) }$
    \STATE $y_{k+1}=\beta_{2} y_{k}+\left(1-\beta_{2}\right)\left(\nabla J\left(\theta_{k}\right)\right)^{2} \text { (Update biased secondary raw moment estimate) }$
    \STATE $\hat{v}_{k+1}=v_{k+1} /\left(1-\beta_{1}^{k+1}\right) \text { (Compute bias-corrected first-order moment estimate) }$
    \STATE $\hat{y}_{k+1}=y_{k+1} /\left(1-\beta_{2}^{k+1}\right) \text { (Compute bias-corrected secondary raw moment estimate) }$
    \STATE $g_{k}=\hat{v}_{k+1} \text{ (Estimate gradients of the objective function) }$
    \STATE $\alpha_{k}=\frac{\alpha} {\sqrt{\hat{y}_{k+1}}+\hat{\epsilon}} \text{ (Adjust effective learning rates) }$
    \STATE $\theta_{k+1}=\theta_{k}-\alpha_{k} g_{k}$
\ENDFOR
\end{algorithmic}
\end{algorithm}

We can compactly write the updating rules by integrating the bias-correction steps into their following steps, thus formulating ADAM in the same format as other algorithms shown in this paper (see Algorithm \ref{algorithm_adam}). Note that the rational factor $\epsilon$ prevents the denominator of the effective learning rate $\alpha_k$ from zero. ADAM has two essential properties, including its moment estimation of gradients and naturally performing self-adaption on learning rates.

\clearpage
\section*{S.V Additional experiments}
\label{appendix_additional_experiments}
\addcontentsline{toc}{section}{Additional experiments}
This section presents additional experimental results not included in the main body due to page limitations.

\setcounter{subsection}{0}
\subsection{MuJoCo tests on TD3}
In addition to using SAC, we also validate RAD on various MoJoCo tasks employing the TD3 algorithm. The results demonstrate consistent superiority of RAD across all tasks (see Fig.~\ref{fig_additional mujoco tasks}), with a notable 9.7\% improvement over ADAM specifically in the Walker2d-v3 task (see Table~\ref{table_tar_td3_mujoco}). These findings highlight the efficacy and versatility of RAD when applied to different RL algorithms.

\begin{figure}[!thbp]
\centering
\subfloat[Ant-v3]{
\label{fig_ant_td3}
\includegraphics[width=0.24\linewidth]{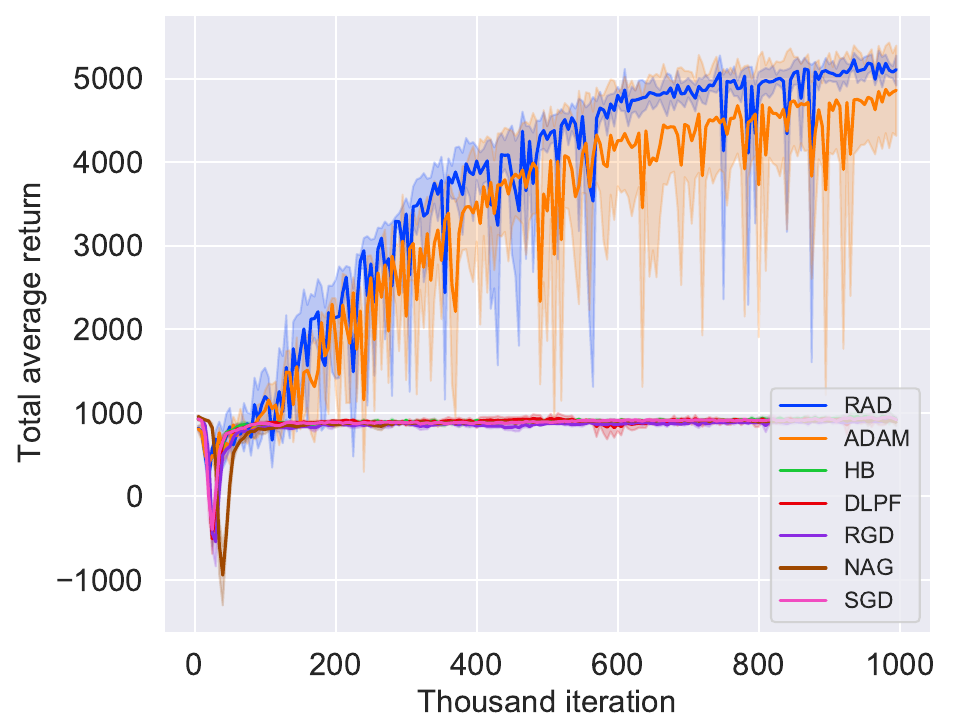}
}
\subfloat[HalfCheetah-v3]{
\label{fig_halfcheetah_td3}
\includegraphics[width=0.24\linewidth]{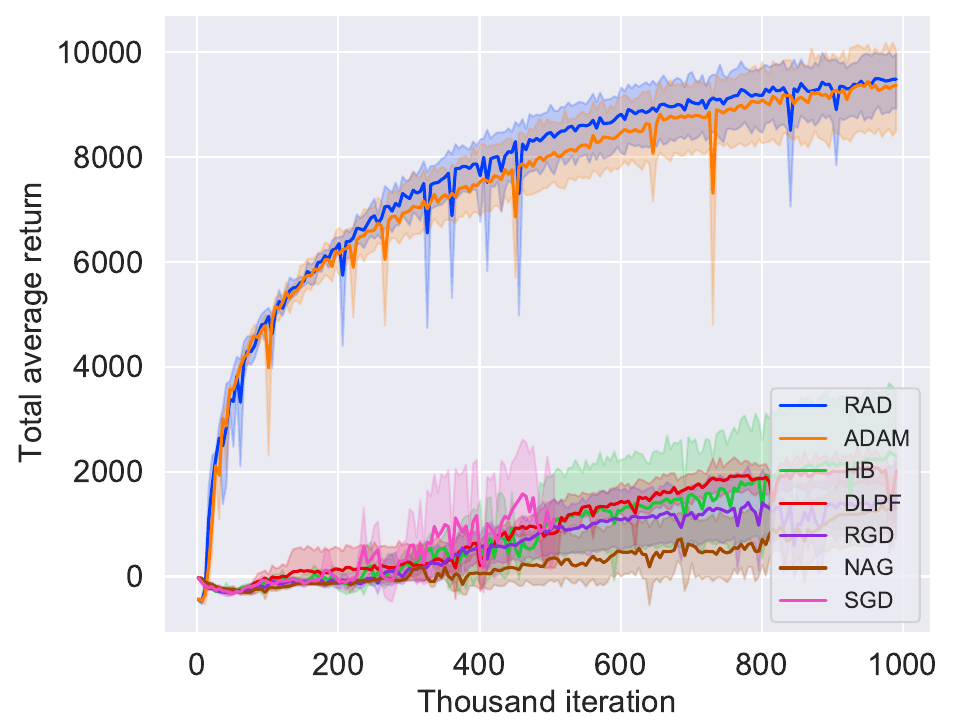}
}
\subfloat[Walker2d-v3]{
\label{fig_walker2d_td3}
\includegraphics[width=0.24\linewidth]{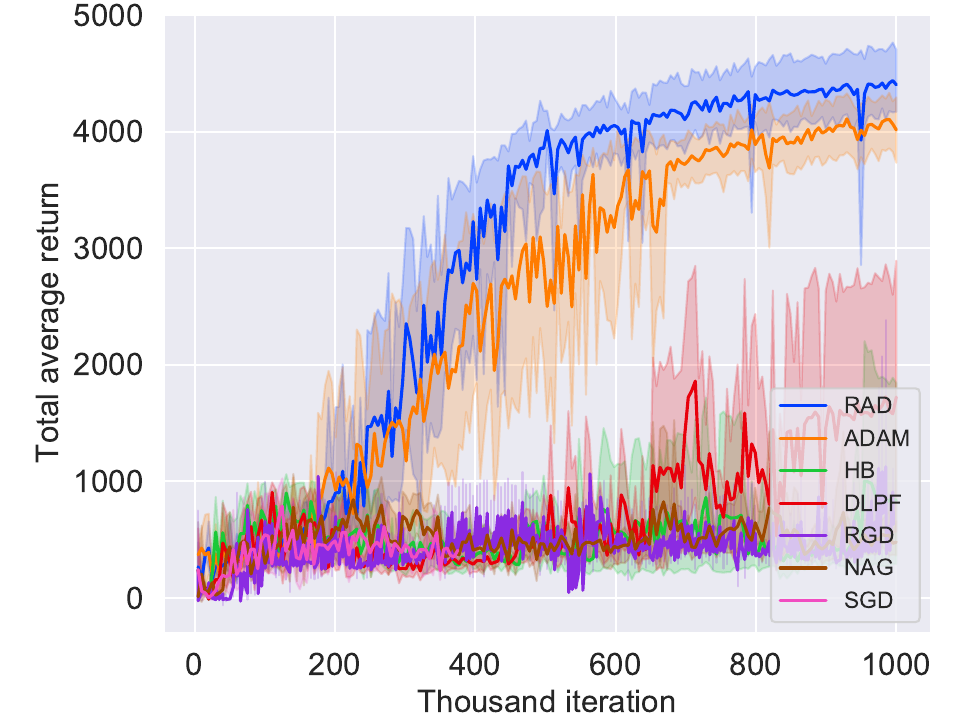}
}
\subfloat[Swimmer-v3]{
\label{fig_swimmer_td3}
\includegraphics[width=0.24\linewidth]{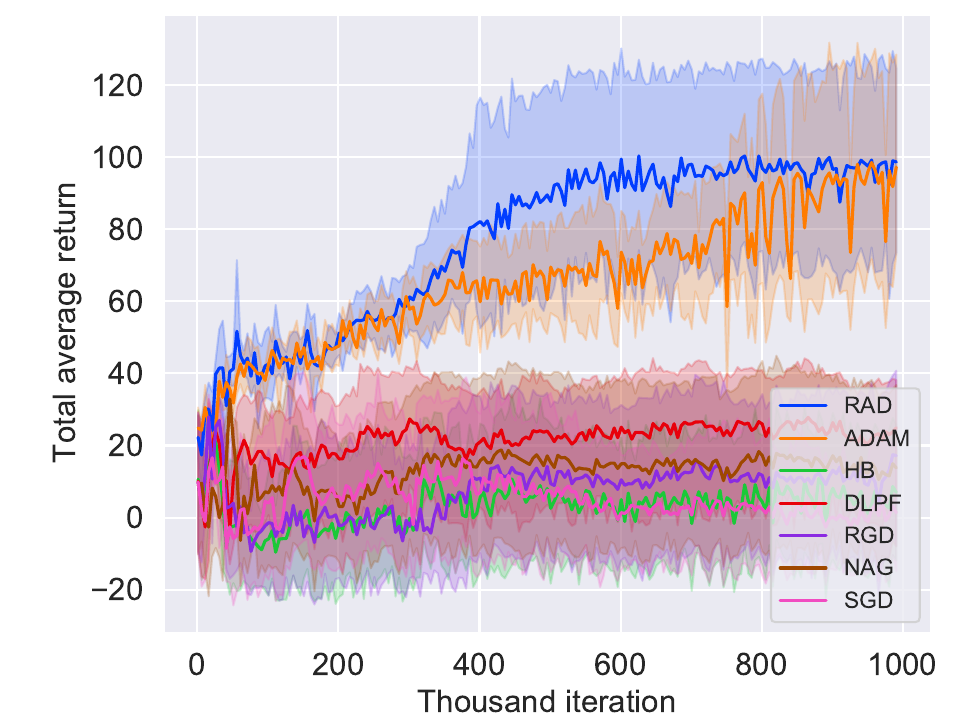}
}
\caption{
\label{fig_additional mujoco tasks} Policy performance of TD3 on MuJoCo tasks. The solid lines correspond to the mean, and the shaded regions correspond to the 95\% confidence interval over five runs.}

\end{figure}

\begin{table}[!tbhp]
  \caption{Policy performance of TD3 on MuJoCo tasks over five runs. The maximum value for each task is bolded. $\pm$ corresponds to the standard deviation. $\rm{DIV}$ represents early divergence. $\Uparrow$ denotes the relative improvement of RAD compared to ADAM.}
  \label{table_tar_td3_mujoco}
  \centering
  \begin{tabular}{lccccc}
    \toprule
    Tasks & Ant-v3 & HalfCheetah-v3 & Walker2d-v3 & Swimmer-v3  \\
    \midrule
    RAD & \textbf{5107}$\pm$205 & \textbf{9559}$\pm$610 & \textbf{4402}$\pm$301 & \textbf{100}$\pm$35 \\
    ADAM & 4860$\pm$647 & 9417$\pm$872 & 4014$\pm$328 & 93$\pm$38 \\
    HB & 924$\pm$19 & 2313$\pm$1465 & 837$\pm$1015 & 2$\pm$22 \\
    DLPF & 917$\pm$16 & 2038$\pm$371 & 1722$\pm$1333 & 24$\pm$24 \\
    RGD & 912$\pm$37 & 1383$\pm$806 & 765$\pm$355 & 15$\pm$29 \\
    NAG & 887$\pm$46 & 1296$\pm$735 & 478$\pm$187 & 11$\pm$26 \\
    SGD & 920$\pm$40 & 2372$\pm$735 & $\rm{DIV}$ & 1$\pm$20 \\
    $\Uparrow$ & 5.1\% & 1.5\% & 9.7\% & 7.5\% \\
    \bottomrule
  \end{tabular}
\end{table}

\subsection{Autonomous driving tasks}
To thoroughly assess RAD's performance in complex real-world tasks, we employ the IDSim benchmark for autonomous driving tests at simulated urban intersections \cite{Jiang2023IDSim}. We generate 20 intersections with varying typologies, each featuring mixed traffic flows comprising cars, bicycles, and pedestrians. To replicate the real-world perception conditions where noise is unavoidable, we deliberately add Gaussian noise to the observations of other road users, as illustrated in Table \ref{tab_noise distri in IDSim}. Additionally, we implement a traffic signal system, where left turns and straight going are controlled by a single signal, creating challenging driving situations such as unprotected left turns. The objective of the task is to maneuver the ego vehicle through intersections by controlling its acceleration and steering angle within a limited number of control steps.

\begin{table}[!tbhp]
  \caption{Standard deviations of observation noise. All noises obey standard Gaussian distributions derived from real-world datasets.}
  \label{tab_noise distri in IDSim}
  \centering
  \begin{tabular}{lccc}
    \toprule
    Observation & Car & Bicycle & Pedestrian  \\
    \midrule
    Longitudinal position (m) & $4.34\times10^{-3}$ & $1.75\times10^{-3}$ & $4.53\times10^{-3}$ \\
    Lateral position (m) & $6.04\times10^{-3}$ & $1.95\times10^{-3}$ & $4.80\times10^{-3}$ \\
    Velocity (m/s) & $7.51\times10^{-2}$ & $2.64\times10^{-2}$ & $6.58\times10^{-2}$ \\
    Heading angle (rad) & $5.32\times10^{-3}$ & $2.17\times10^{-3}$ & $3.65\times10^{-2}$\\
    \bottomrule
  \end{tabular}
\end{table}

We compare RAD with the widely-used ADAM optimizer within the context of the ADP algorithm \cite{Powell2011ADP}. After convergence, we test the two policies at each intersection 50 times to assess their driving performance. Table~\ref{tab.diving performance} presents five evaluation metrics: the success rate, collision rate, failure rate, driving comfort, and travel efficiency. The success rate represents the ratio of successful navigation through the intersection, while the collision rate measures instances of collisions with other road users. The failure rate encompasses cases where the ego vehicle either goes out of the drivable area or exceeds the maximum control steps. Driving comfort is evaluated by the root mean square of acceleration and yaw rate, and travel efficiency denotes the relative ratio between ego speed and road speed limits.

The results demonstrate that the policy trained with RAD achieves a remarkable success rate of 93.2\%, significantly outperforming ADAM by a margin of 3.7\%. Specifically, The policy trained with RAD exhibits lower collision rates, improved driving comfort, and higher travel efficiency than ADAM. Due to insufficient long-term training stability, the policy trained with ADAM experiences difficulties in accurately interpreting complex traffic conditions at intersections, leading to inadequate control commands and unsatisfactory interactions with other traffic participants. On the other hand, RAD stabilizes the training process by reducing the effects of noisy policy gradients arising from the intricate driving environment. Consequently, RAD produces more effective and reliable driving policies for autonomous vehicles. This experiment validates that RAD's applicability extends beyond standard RL benchmarks, such as MuJoCo and Atari, to real-world applications like autonomous driving. It highlights the potential of RAD as a robust optimization algorithm for complex, real-world tasks. Additionally, it is essential to note that ADP represents a model-based RL algorithm, which significantly diverges from the model-free nature of previously employed algorithms such as DDPG\cite{lillicrap2016continuous}, SAC\cite{haarnoja2018soft}, TD3\cite{fujimoto2018addressing}, and DQN\cite{mnih2015human}. This further underscores the scalability of RAD across various RL settings.

\begin{table}[!tbhp]
\centering
\caption{Driving performance at urban intersections. Better metrics are highlighted in bold.}
\label{tab.diving performance}
\begin{tabular}{cccccc}
\toprule
{Optimizer} & Success rate$\uparrow$ & Collision rate$\downarrow$ & Failure rate$\downarrow$ & Driving comfort$\downarrow$ & Travel efficiency$\uparrow$ \\
\midrule
RAD & \textbf{93.2\%} & \textbf{2.9\%} & \textbf{3.9\%} & \textbf{0.78} & \textbf{0.73} \\
ADAM & 89.5\% & 5.8\% & 4.7\% & 1.00 & 0.68 \\
\bottomrule
\end{tabular}
\end{table}

\section*{S.IV Experimental settings}
\label{appendix_experimental_settings}
\addcontentsline{toc}{section}{Experimental settings}
The detailed experimental settings are shown in Table \ref{table_experimental settings I} and Table \ref{table_experimental settings II}.

\begin{table}[hp]
  \caption{Experimental settings I}
  \label{table_experimental settings I}
  \centering
  \begin{tabular}{llll}
    \toprule
    Task & CartPole-v1 & Hopper-v3 & Other MuJoCo tasks \\
    \midrule
    RL algorithm & {DDPG \& SAC} & SAC & SAC \\
    Discount factor & 0.99 & 0.99 & 0.99\\
    Exploration noise & $\varepsilon\sim\mathcal{N}(0,0.1)$ & / & / \\
    Temperature coefficient & 0.2 & 0.2 & 0.2\\
    Approximate function & MLP & MLP & MLP \\
    FC layer size & $256\times256$ & $256\times256$ & $256\times256$ \\
    Activation function & ReLU & ReLU & ReLU \\
    Learning rate decay & / & CosineAnnealingLR & /\\
    Critic learning rate & $5\times10^{-4}$ & $1\times10^{-3}\rightarrow1\times10^{-4}$ & $1\times10^{-3}$\\
    Actor learning rate & $5\times10^{-5}$ & $1\times10^{-3}\rightarrow1\times10^{-4}$ & $1\times10^{-3}$ \\
    Target network learning rate & $5\times10^{-3}$ & $5\times10^{-3}$ & $5\times10^{-3}$ \\
    Maximum iteration & $3\times10^4$ & $5\times10^5$ & $1\times10^6$\\
    Batch size & 1024 & 256 & 256\\
    First-order momentum coeff. & 0.9 & 0.9 & 0.9 \\
    Second-order momentum coeff. & 0.999 & 0.999 & 0.999 \\
    
    \bottomrule
  \end{tabular}
\end{table}

\begin{table}[hp]
  \caption{Experimental settings II}
  \label{table_experimental settings II}
  \centering
  \begin{tabular}{lllll}
    \toprule
    Task & MuJoCo tasks & Atari games & IDSim \\
    \midrule
    RL algorithm & TD3 & DQN & ADP\\
    Discount factor & 0.99 & 0.99 & 0.95 \\
    Exploration noise & $\varepsilon\sim\mathcal{N}(0,0.1)$ & $\varepsilon$-greedy policy & / \\
    Approximate function & MLP & CNN & MLP \\
    Number of conv. layers & / & 3 & / \\
    FC layer size & $256\times256$ & $512$ & $256^5$ \\
    Activation function & ReLU & ReLU & ReLU \\
    Learning rate decay & / & / & LinearLR \\
    Critic learning rate & $3\times10^{-4}$ & $1\times10^{-4}$ & $2\times10^{-5} \rightarrow 0$\\
    Actor learning rate & $3\times10^{-4}$ & / & $2\times10^{-5} \rightarrow 0$ \\
    Target network learning rate & $5\times10^{-3}$ & / & / \\
    Maximum iteration &  $1\times10^6$ & $1\times10^6$ & $3\times10^5$\\
    Batch size & 32 & 256 & 256\\
    First-order momentum coeff. & 0.9 & 0.9 & 0.9 \\
    Second-order momentum coeff. & 0.999 & 0.999 & 0.999 \\
    
    \bottomrule
  \end{tabular}
\end{table}

\end{onecolumn}
\end{document}